\documentclass[10pt,english,english,10pt,journal,english,compsoc]{IEEEtran}
\usepackage[LGR,T1]{fontenc}

\usepackage[latin9]{inputenc}
\usepackage{color}
\usepackage{babel}
\usepackage{float}
\usepackage{textcomp}
\usepackage{amsmath}
\usepackage{amssymb}
\usepackage{stmaryrd}
\usepackage{graphicx}
\usepackage{hyperref}

\newcommand{\lyxmathsym}[1]{\ifmmode\begingroup\def\b@ld{bold}
  \text{\ifx\math@version\b@ld\bfseries\fi#1}\endgroup\else#1\fi}

\providecommand{\tabularnewline}{\\}
\floatstyle{ruled}
\newfloat{algorithm}{tbp}{loa}
\providecommand{\algorithmname}{Algorithm}
\floatname{algorithm}{\protect\algorithmname}

\DeclareRobustCommand{\greektext}{%
  \fontencoding{LGR}\selectfont\def\encodingdefault{LGR}}
\DeclareRobustCommand{\textgreek}[1]{\leavevmode{\greektext #1}}
\DeclareFontEncoding{LGR}{}{}
\DeclareTextSymbol{\~}{LGR}{126}
\providecommand{\tabularnewline}{\\}

\providecommand{\tabularnewline}{\\}
\floatstyle{ruled}
\newfloat{algorithm}{tbp}{loa}
\providecommand{\algorithmname}{Algorithm}
\floatname{algorithm}{\protect\algorithmname}

\def\U{\mathcal{U}}

\definecolor{gray}{rgb}{0.5,0.5,0.5}

\usepackage{babel}

\usepackage{url}

\usepackage{color}
\newcommand*{\red}{\textcolor{black}} 

\usepackage{array}
\usepackage{multirow}

\makeatother

\begin{document}
\title{Robust  Subjective Visual Property Prediction from Crowdsourced Pairwise Labels} 

\author{Yanwei~Fu, Timothy~M.~Hospedales, Tao~Xiang, Jiechao~Xiong,\\
  Shaogang~Gong,   Yizhou~Wang, and Yuan~Yao
\IEEEcompsocitemizethanks{
 \IEEEcompsocthanksitem  Yanwei~Fu, Timothy~M.~Hospedales, Tao~Xiang,   and Shaogang~Gong are with the School of Electronic Engineering and Computer Science, Queen Mary University of London, E1 4NS, UK. Email: \{y.fu, t.hospedales,t.xiang,s.gong\}@qmul.ac.uk. 
\IEEEcompsocthanksitem Jiechao~Xiong, and Yuan Yao  are with  the School of Mathematical Sciences, Peking University, China. Email: xiongjiechao@pku.edu.cn, yuany@math.pku.edu.cn.  Yizhou Wang is with Nat'l Engineering Laboratory for Video Technology
Cooperative Medianet Innovation Center Key Laboratory of Machine Perception (MoE) Sch'l of EECS, Peking University, Beijing, 100871, China. Email: yizhou.wang@pku.edu.cn. 

} 
\thanks{}
}
\IEEEcompsoctitleabstractindextext{ 
\begin{abstract} 
The problem of estimating subjective visual properties from image and video has attracted increasing interest. A subjective visual property is  useful either on its own (e.g.~image and video interestingness) or as an intermediate representation for visual recognition (e.g.~a relative attribute). Due to its ambiguous nature,  annotating the value of a subjective visual property for learning a prediction model is challenging.  To make the annotation  more reliable, recent studies employ crowdsourcing tools to collect pairwise comparison labels. However, using crowdsourced data also introduces outliers. Existing methods rely on majority voting to prune the annotation outliers/errors. They thus require a large amount of pairwise labels to be collected. More importantly as a local outlier detection method, majority voting is ineffective in identifying outliers that can cause global ranking inconsistencies.  In this paper, we propose a more principled way to identify annotation outliers by formulating the subjective visual property prediction task as a unified robust learning to rank problem, tackling both the outlier detection and learning to rank jointly. This differs from existing methods in that (1) the proposed method integrates local pairwise comparison labels together to minimise a cost that corresponds to global inconsistency of ranking order, and (2)  the outlier detection and learning to rank problems are solved jointly. This not only leads to  better detection of annotation outliers but also enables learning with extremely sparse annotations. 

\end{abstract} 
}
\maketitle

\section{Introduction}


The solutions to many computer vision problems involve the estimation of some visual properties of an image or video, represented as either discrete or continuous variables. For example scene classification  aims to estimate the value of a discrete variable indicating which scene category an image belongs to; for object detection the task is to estimate a binary variable corresponding the presence/absence of the object of interest and a set of variables indicating its whereabouts in the image plane (e.g.~four variables if the whereabouts are represented as bounding boxes).
Most of these visual properties are objective; that is,  there is no or little ambiguity in their true values to a human annotator. 

In comparison, the problem of  estimating subjective visual properties is much less studied. This class of computer vision problems nevertheless encompass a variety of important applications. For example: estimating attractiveness \cite{grauman2011rationales} from faces would interest social media or online dating websites; and estimating  properties of consumer goods such as  shininess of shoes \cite{whittlesearch} improves customer experiences on online shopping websites.  Recently, the problem of automatically predicting if people would find an image or video interesting has started to receive increasing attention \cite{Dhar2011cvpr,imginterestingnessICCV2013,yugangVideoInteresting2013}. Interestingness prediction has a number of real-world applications. In particular, since the number of images and videos uploaded to the Internet is growing explosively, people are increasingly relying on image/video  recommendation tools to select which ones to view. Given a query, ranking the retrieved data with relevance to the query based on the predicted interestingness would improve  user satisfaction. Similarly user stickiness can be increased if a media-sharing website such as YouTube can recommend videos that are both relevant and interesting.  Other applications such as web advertising and video summarisation can also benefit. Subjective visual properties such as the above-mentioned ones are useful on their own. But they can also be used as an intermediate representation for other tasks such as visual recognition, e.g., different people can be recognised by how pale their skin complexions are and how chubby their faces are \cite{parikh2011relativeattrib}. When used as a semantically meaningful representation, these subjective visual properties often are referred to as relative attributes  \cite{whittlesearch,parikh2011relativeattrib,robust_relative_attrib}.


Learning a model for subjective visual property (SVP) prediction is challenging primarily due to the difficulties in obtaining annotated training data. Specifically, since most SVPs can be represented as continuous variables (e.g.~an interestingness/aesthetics/shininess score with a value range of 0 to 1 with 1 being most interesting/aesthetically appealing/shinning),  SVP prediction  can be cast as a regression problem -- the low-level feature values are regressed to the SVP values given a set of training data annotated with their true SVP values. However, since by definition these properties are subjective, different human annotators often struggle to give an absolute value and as a result the annotations of different people on the same instance can vary hugely.  For example, on a scale of 1 to 10,  different people will have very different ideas on what a scale 5 means for an image, especially without any common reference point. On the other hand, it is noted that humans can in general more accurately rank a pair of data points in terms of their visual properties \red{~\cite{Chen2009TSR,age_ranking}} , e.g.~it is easier to judge which of  two images is  more interesting relatively than giving an absolute interestingness score to each of them. Most existing studies \cite{whittlesearch,grauman2011rationales,age_ranking} on SVP prediction thus take a  learning to rank  approach  \cite{chapelle2010efficientRankSVM}, where annotators give comparative labels about pairs of images/videos  and the learned model is a ranking function that predicts the SVP value as a ranking score.



\begin{figure}
\centering{}\includegraphics[scale=0.55]{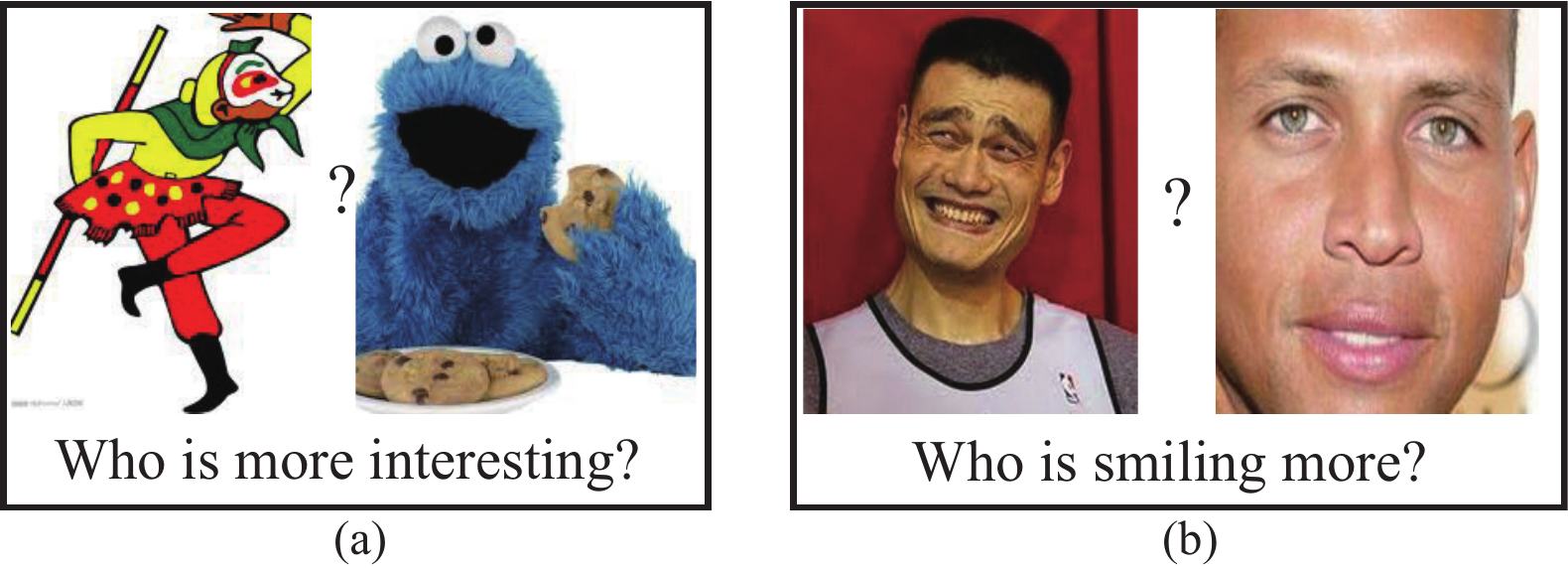}
\caption{\label{fig:Relative-attributes-are}Examples of pairwise comparisons of subjective visual properties.  }
\end{figure}

\red{To annotate these pairwise comparisons, crowdsourcing tools such as Amazon Mechanic Turk (AMT) are resorted to, which allow a large number of annotators to collaborate at very low cost. Data annotation based on crowdsourcing is increasingly popular \cite{parikh2011relativeattrib,whittlesearch,imginterestingnessICCV2013,yugangVideoInteresting2013} recently for annotating large-scale datasets. } However, this brings about two new problems: (1) \textbf{Outliers} -- The crowd is not all trustworthy: it is well known that crowdsourced data are greatly affected by noise and outliers~\cite{chen13WSDM,wu2011IJCAI,Ganghua2013iccv} which can be caused by a number of factors. Some workers may be lazy or malicious \cite{crowduser}, providing random or wrong annotations either carelessly or intentionally; some other outliers are unintentional human errors caused by the ambiguous nature of the data,  thus are unavoidable regardless how good the attitudes of the workers are. For example, the pairwise ranking  for Figure \ref{fig:Relative-attributes-are}(a) depends  on the cultural/psychological background of the annotator -- whether s/he is more familiar/prefers  the story of  Monkey King or  Cookie Monster\red{\footnote{\red{This is also known as Halo Effect in Psychology.}}}. \red{ When we learn the model from labels collected from many people, we essentially aim to learn the consensus, i.e.~what most people would agree on. Therefore, if most of the annotators growing up watching Sesame Street thus consciously or subconsciously consider the Cookie Monster to be more interesting than the Monkey King, their pairwise labels/votes would represent the consensus. In contrast, one annotator who is familiar with the stories in Journey to the West may choose the opposite; his/her label is thus an outlier under the consensus.} 
(2) \textbf{Sparsity} -- the number of pairwise comparisons required is much bigger than the number of data points because $n$ instances define a $\mathcal{O}(n^{2})$ pairwise space. Consequently, even with crowdsourcing tools, the annotation remains be sparse, i.e.~not all pairs are compared and each pair is only compared a few times.

To deal with  the outlier problem in crowdsourced data, existing studies take a \red{ majority voting strategy \cite{parikh2011relativeattrib,whittlesearch,imginterestingnessICCV2013,Kovashka_2013_ICCV,conf/nips/WelinderBBP10, Raykar:2009:SLM:1553374.1553488, Whose_vote}. } That is, a large budget of $5-10$ times the number of actual annotated pairs required is allocated to obtain multiple annotations for each pair. These annotations are then averaged over so as to eliminate label noise. However, the effectiveness of the majority voting strategy is often limited by the sparsity problem -- it is typically infeasible to have many annotators for each pair. Furthermore, there is no guarantee that outliers, particularly those caused by unintentional human errors can be dealt with effectively. This is because majority voting is a local consistency detection based strategy -- when there are contradictory/inconsistent pairwise rankings for a given pair, the pairwise rankings receiving minority votes are eliminated as outliers. However, it has been found that when pairwise local rankings are integrated into a global ranking, it is possible to detect outliers that can cause global inconsistency and yet are locally consistent, i.e.~supported by majority votes \cite{Hodgerank}. Critically,  outliers that cause global inconsistency have more significant detrimental effects on learning a ranking function for  SVP prediction and thus should be the main focus of an outlier detection method.


In this paper we propose a novel approach to subjective visual property prediction from sparse and noisy pairwise comparison labels collected using crowdsourcing tools. Different from existing approaches which first remove outliers by majority voting, followed by regression \cite{imginterestingnessICCV2013} or learning to rank \cite{yugangVideoInteresting2013}, we formulate a unified robust learning to rank (URLR) framework to  solve jointly both the outlier detection and learning to rank problems. Critically, instead of detecting outliers locally and independently  at each pair by majority voting,  our outlier detection method operates globally, integrating all local pairwise comparisons together to minimise a cost that corresponds to global inconsistency of ranking order. This enables us to identify those outliers that receive majority votes but cause large global ranking inconsistency and thus should be removed. Furthermore, as a global method that aggregates comparisons across different pairs, our method can operate with as few as one comparison per pair, making our method much more robust against the data sparsity problem compared to the conventional majority voting approach that aggregates comparisons for each pair in isolation. More specifically, the proposed model generalises a partially
penalised LASSO optimisation or Huber-LASSO formulation \cite{lasso,Gannaz07,NIPS2013_5104} from a robust statistical ranking formulation to a robust learning to rank model, making it suitable for SVP prediction given unseen images/videos. We also formulate a regularisation path based solution to solve this new formulation efficiently.   Extensive experiments are carried out on benchmark datasets including two image and video interestingness datasets \cite{imginterestingnessICCV2013,yugangVideoInteresting2013} and two relative attribute datasets \cite{whittlesearch}. The results demonstrate that our method significantly outperforms the state-of-the-art alternatives. 

\vspace{-0.3cm}
\section{Related work}

\noindent \textbf{Subjective visual properties}\quad{} Subjective visual property prediction covers a large variety of computer vision problems; it is thus beyond the scope of this paper to present an exhaustive review here. Instead we focus mainly on the image/video interestingness prediction problem which share many characteristics with other SVP prediction problem such as image quality~\cite{keimgquality2006CVPR}, memorability~\cite{Isola2011cvpr},  
and aesthetics~\cite{Dhar2011cvpr} prediction.

\vspace{0.1cm}
\noindent \textbf{Predicting image and video interestingness}\quad Early efforts on image interestingness prediction focus on different aspects than interestingness as such, including    memorability \cite{Isola2011cvpr}
and aesthetics \cite{Dhar2011cvpr}. These SVPs are related to interestingness but different. For instance, it is found that memorability can have a low correlation with interestingness - people often remember things that they find uninteresting \cite{imginterestingnessICCV2013}. The work of Gygli et al \cite{imginterestingnessICCV2013} is the first systematic study of image interestingness. It shows that three cues contribute the most to interestingness: aesthetics, unusualness/novelty and general preferences, the last of which refers to the fact that people in general find certain types of scenes more interesting than others, for example outdoor-natural vs.~indoor-manmade. Different features are then designed to represent these cues as input to a prediction model. In comparison, video interestingness has received much less attention, perhaps because it is even harder to understand its meaning and contributing cues. Liu et al.~\cite{Liu2009IJCAI} focus on key frames so essentially treats it as an image interestingness problem, whilst \cite{yugangVideoInteresting2013} is the first work that proposes benchmark video interestingness datasets and evaluates different features for video interestingness prediction. 

Most earlier works cast the aesthetics or interestingness prediction problem as a regression problem \cite{keimgquality2006CVPR,Dhar2011cvpr,Isola2011cvpr,Liu2009IJCAI}. However, as discussed before, obtaining an absolute value of interestingness for each data point is too subjective and affected too much by unknown personal preference/social background to be reliable. Therefore the most recent two studies on image \cite{imginterestingnessICCV2013} and video \cite{yugangVideoInteresting2013} interestingness  all  collect pairwise comparison data by crowdsourcing. Both use majority voting to remove outliers first. After that the prediction models differ -- \cite{imginterestingnessICCV2013} converts pairwise comparisons into an absolute interestingness values and use a regression model, whilst \cite{yugangVideoInteresting2013} employs rankSVM \cite{chapelle2010efficientRankSVM} to learn a ranking function, with the estimated ranking score of an unseen video used as the interestingness prediction. We compare with both approaches in our experiments and demonstrate that our unified robust learning to rank approach is superior as we can remove  outliers more effectively -- even if they correspond to comparisons receiving majority votes, thanks to its global formulation.

\vspace{0.1cm}
\noindent \textbf{Relative attributes}\quad In a broader sense interestingness can be considered as one type of relative attribute \cite{parikh2011relativeattrib}. Attribute-based modelling \cite{lampert2009zeroshot_dat,farhadi2009attrib_describe} has gained popularity recently as a way to \emph{describe} instances and classes at an intermediate level of representation. Attributes are then used for various tasks including N-shot and zero-shot transfer learning. Most previous studies consider binary attributes  \cite{lampert2009zeroshot_dat,farhadi2009attrib_describe}. Relative attributes \cite{parikh2011relativeattrib}
were recently proposed to learn a ranking function to predict relative semantic strength of visual attributes. Instead of the
original class-level attribute comparisons   in \cite{parikh2011relativeattrib},
this paper focuses on instance-level comparisons due to the huge intra-class
variations in real-world problems. With instance-level pairwise comparisons,
relative attributes have been used for interactive image search \cite{whittlesearch},
and semi-supervised \cite{ShrivastavaECCV12} or active learning \cite{attrclsfier,BiswasCVPR13}
of visual categories. However, no previous work addresses
the problem of annotation outliers except \cite{whittlesearch}, which adopts the heuristic majority voting strategy.

\vspace{0.1cm}
\noindent \textbf{Learning from noisy paired crowdsourced data}\quad{}Many
large-scale computer vision problems rely on human intelligence tasks
(HIT) using crowdsourcing services, e.g.~AMT (Amazon Mechanical Turk)
to collect annotations. Many studies \cite{crowduser,amazon_mechanical,SUN_attrib,Ganghua2013iccv}
highlight the necessity of validating the random or malicious labels/workers
and give filtering heuristics for data cleaning. However, these are primarily based on majority voting
which requires a costly volume of redundant annotations, and has
no theoretical guarantee of solving the outlier and sparsity problems.
As a local (per-pair) filtering method, majority voting does not respect global ordering and even risks introducing additional
inconsistency due to the well-known Condorcet's paradox in social choice and voting theory \cite{condorcet}.
\red{  Active learning \cite{active_relative,attrclsfier,BiswasCVPR13} is  an another way to circumvent the  $\mathcal{O}(n^{2})$ pairwise labelling space. It actively poses specific requests to annotators and learns from their feedback, rather than the `general' pairwise comparisons discussed in this work. }
\red{  
 Besides paired crowdsourced data, majority voting is more widely used in crowdsourcing where multiple annotators directly label instances, which attracted lots of attention in the machine learning community \cite{conf/nips/WelinderBBP10, Raykar:2009:SLM:1553374.1553488, Whose_vote, Kovashka_2013_ICCV}. In contrast, our work focuses on pairwise comparisons which are relatively easier for annotators in evaluating the subjective visual properties~\cite{Chen2009TSR} . }

\vspace{0.1cm}
\noindent \textbf{Statistical ranking and learning to rank} Statistical ranking has been widely studied in
statistics and computer science
\cite{video_quality_hodgerank,Online_hodge,Chen2009TSR,angularembeddingobjseg}.
However, statistical ranking only concerns the ranking of the observed/training data, but not learning to predict unseen data by learning ranking functions. To learn  ranking functions for applications such as interestingness prediction, a feature representation of the data points must be used as model input in addition to the local ranking orders. This is addressed in learning to rank which is widely studied in machine learning \cite{pairwiserank,browserank,robust_ranking_learning}.
However, existing learning to rank works do not explicitly model and remove outliers for robust learning: a critical issue for learning from crowdsourced data in practice. In this work, for the first time, we study the problem of robust learning to rank given extremely noisy and sparse crowdsourced pairwise labels. We show both theoretically and experimentally  that by solving both the outlier detection and ranking prediction problems jointly, we achieve better outlier detection than existing statistical ranking methods and better ranking prediction than existing learning to rank method such as RankSVM without outlier detection.

\vspace{0.1cm}
\noindent \textbf{Our contributions} are threefold: (1) We propose
a novel robust learning to rank method for subjective visual property prediction
using noisy and sparse pairwise comparison/ranking labels as training data. (2) For the first
time, the problems of detecting outliers and estimating linear ranking models
are solved jointly in a unified framework. (3) We demonstrate both
theoretically and experimentally that our method is superior to existing
majority voting based methods as well as statistical ranking based
methods. An earlier and preliminary version of this work is presented in \cite{interestingnessECCV2014}
which focused only on the image/video interestingness prediction problem.

\vspace{-0.3cm}
\section{Unified Robust Learning to Rank}

\subsection{Problem definition}

We aim to learn a subjective visual property (SVP) prediction model
from a set of sparse and noisy pairwise comparison labels, each comparison
corresponding to a local ranking between a pair of images or videos.
Suppose our training set has $N$ data points/instances represented by a feature matrix $\Phi=\left[\mathbf{\boldsymbol{\phi}}_{i}^{T}\right]_{i=1}^{N}\in\mathbb{R}^{N\times d}$,
where $\mathbf{\boldsymbol{\phi}}_{i}$ is a $d$-dimensional column low-level
feature vector representing instance $i$. The pairwise comparison labels (annotations collected using crowdsourcing tools) can be naturally represented as a directed comparison graph   \textbf{$G=\left(V,E\right)$}
with a node set 
$V=\left\{ i\right\} _{i=1}^{N}$ corresponding to the $N$ instances and an edge set $E=\{e_{ij}\}$ corresponding to the pairwise comparisons.

The pairwise comparison labels can be provided by multiple annotators.  They are dichotomously
saved: Suppose  annotator $\alpha$ gives a pairwise comparison for
instance $i$ and $j$ ($i,j\in V$). If $\alpha$ considers that the  SVP of instance $i$ is stronger/more than that  of $j$,
we save $(i,\, j,\, y_{e_{ij}}^{\alpha})$ and set $y_{e_{ij}}^{\alpha}=1$.
If the opposite is the case, we save $(j,\, i,\, y_{e_{ji}}^{\alpha})$
and set $y_{e_{ji}}^{\alpha}=1$.  All the pairwise comparisons between instances
$i$ and $j$ are then aggregated over all annotators who have cast a vote on this pair; the results are represented as  $w_{e_{ij}}=\sum_{\alpha}\llbracket y_{e_{ij}}^{\alpha}=1\rrbracket$ which 
is the total number of votes on $i$ over $j$ for a specific SVP, where $\llbracket\rrbracket$
indicates the Iverson\textquoteright s bracket notation, and $w_{e_{ji}}$ which is defined similarly. This gives an edge weight vector $\boldsymbol{w}=\left[w_{e_{ij}}\right]\in\mathbb{R}^{\mid E\mid}$ where $|E|$ is the number of edges. Now the edge set can be represented as $E=\left\{ e_{ij}|w_{e_{ij}}>0\right\} $
and $w_{e_{ij}}\in\mathbb{R}$ is the weight for the
edge $e_{ij}$. In other words, an edge $e_{ij}$: $i\rightarrow j$
exists if $w_{e_{ij}}>0$. The topology of the graph is denoted by a flag indicator vector
 $\boldsymbol{y}=\left[y_{e_{ij}}\right]\in\mathbb{R}^{\mid E\mid}$ where each  indicator $y_{e_{ij}}=1$ indicates that there is an edge between instances $i$ to $j$ regardless how many votes it carries. Note that all the elements in $\boldsymbol{y}$ have the value $1$, and their index ${e_{ij}}$ gives the corresponding nodes in the graph.

Given the training data consisting of the feature matrix $\Phi$ and the annotation graph $G$, there are two tasks:
\begin{enumerate}
\item Detecting and removing the outliers in the edge set $E$ of $G$. To this end, we introduce a set of unknown variables $\boldsymbol{\gamma}=\left[\gamma_{e_{ij}}\right]\in\mathbb{R}^{\mid E\mid}$ where each variable $\gamma_{e_{ij}}$ indicates  
whether the edge $e_{ij}$
is an outlier. The outlier detection problem thus becomes the problem of estimating $\boldsymbol{\gamma}$.
\item Estimating a prediction function for SVP. In this work a linear model
is considered due to its low computational complexity, that is, given
the low-level feature $\boldsymbol{\phi}_{x}$ of a test instance
$x$ we use a linear function $f(x)=\boldsymbol{\beta}^{T}\boldsymbol{\phi}_{x}$
to predict its SVP, where $\boldsymbol{\beta}$ is the coefficient
weight vector of the low-level feature $\boldsymbol{\phi}_{x}$. Note that all
formulations can be easily updated to use a non-linear function.
\end{enumerate}

So far in the introduced notations three vectors share indices: the flag indicator vector $\boldsymbol{y}$, the outlier variable vector $\boldsymbol{\gamma}$
and the edge weight vector $\boldsymbol{w}$. For notation convenience,
from now on we use $y_{ij}$, $\gamma_{ij}$ and $w_{ij}$
to replace $y_{e_{ij}}$, $\gamma_{e_{ij}}$
and $w_{e_{ij}}$ respectively. As in most graph based model formulations, we define $C\in\mathbb{R}^{\mid E\mid\times N}$
as the incident matrix of the directed graph $G$, where $C_{e_{ij}i}=-1/1$
if the edge $e_{ij}$ enters/leaves vertex $i$. 

Note that in an ideal case, one hopes that the votes received on each pair are unanimous, e.g.~$w_{ij}>0$ and $w_{ji}=0$; but often there are disagreements, i.e.~we have both $w_{ij}>0$ and $w_{ji}>0$. Assuming both cannot be true simultaneously, one of them will be an outlier. In this case, one is the majority and the other minority which will be pruned by the majority voting method. This is why majority voting is a local outlier detection method and requires as many votes per pair as possible to be effective (the wisdom of a crowd).

\vspace{-0.3cm}
\subsection{Framework formulation}
In contrast to majority voting, we propose to prune outliers globally and jointly with learning the SVP prediction function. To this end, the outlier variables $\gamma_{ij}$ for outlier detection and the coefficient weight vector $\boldsymbol{\beta}$
for SVP prediction are estimated in a unified framework. Specifically, for each
edge $e_{ij}\in E$, its corresponding flag indicator $y_{ij}$ is modelled as 
\begin{equation}
y_{ij}=\boldsymbol{\beta}^{T}\boldsymbol{\phi}_{i}-
\boldsymbol{\beta}^{T}\boldsymbol{\phi}_{j}+\gamma_{ij}+\varepsilon_{ij}\label{eq:Y_ij},
\end{equation}
where  $\varepsilon_{ij}\sim\mathcal{N}(0,\sigma^{2})$ is the Gaussian noise 
with zero mean and a variance $\sigma$, and the outlier variable $\gamma_{ij}\in \mathbb{R}$ is assumed to have a higher magnitude than $\sigma$.
 For an edge $e_{ij}$, if $y_{ij}$ is
not an outlier, we expect $\boldsymbol{\beta}^{T}\boldsymbol{\phi}_{i}-\boldsymbol{\beta}^{T}\boldsymbol{\phi}_{j}$
should be approximately equal to $y_{ij}$, therefore
we have $\gamma_{ij}=0$. On the contrary, when the prediction
of $\boldsymbol{\beta}^{T}\boldsymbol{\phi}_{i}-\boldsymbol{\beta}^{T}\boldsymbol{\phi}_{j}$
differs greatly from $y_{ij}$, we can explain $y_{ij}$
as an outlier and compensate for the discrepancy between the prediction
and the annotation with a nonzero value of $\gamma_{ij}$.
The only prior knowledge we have on $\gamma_{ij}$ is
that it is a sparse variable, i.e.~in most cases $\gamma_{ij}=0$.

For the whole training set, Eq~(\ref{eq:Y_ij}) can be re-written in its
matrix form 
\begin{align}
\boldsymbol{y} & =C\Phi\boldsymbol{\beta}+\boldsymbol{\gamma}+\boldsymbol{\epsilon}\label{eq:regression_basic_outlier}
\end{align}
where $\boldsymbol{y}=\left[y_{ij}\right]\in\mathbb{R}^{\mid E\mid}$,
$\boldsymbol{\gamma}=\left[\gamma_{ij}\right]\in\mathbb{R}^{\mid E\mid}$,
$\boldsymbol{\epsilon}=\left[\varepsilon_{ij}\right]\in\mathbb{R}^{\mid E\mid}$
and $C\in\mathbb{R}^{\mid E\mid\times N}$ is the incident matrix
of the annotation graph $G$.

In order to estimate the $ |E|+d$
unknown parameters ($|E|$ for $\boldsymbol{\gamma}$ and $d$ for $\boldsymbol{\beta}$), we
aim to minimise the discrepancy between the annotation $\boldsymbol{y}$ and our
prediction $C\Phi\boldsymbol{\beta}+\boldsymbol{\gamma}$, as well as keeping the outlier estimation
$\boldsymbol{\gamma}$ sparse. Note that $\boldsymbol{y}$ only contains information about which pairs of instances have received votes, but not how many. The discrepancy thus needs to weighted by the number of votes received, represented by the edge weight vector $\boldsymbol{w}=\left[w_{ij}\right]\in\mathbb{R}^{\mid E\mid}$. To that end, we put a weighted $l_{2}-$loss on
the discrepancy and a sparsity enhancing penalty on the outlier variables. This gives us the following cost function: 
\begin{equation}
L(\boldsymbol{\beta},\boldsymbol{\gamma})=\frac{1}{2}\|\boldsymbol{y}-C\Phi\boldsymbol{\beta}-\boldsymbol{\gamma}\|_{2,\boldsymbol{w}}^{2}+p_{\lambda}(\boldsymbol{\gamma})\label{eq:HLasso_general}
\end{equation}
where 
\[
\|\boldsymbol{y}-C\Phi\boldsymbol{\beta}-\boldsymbol{\gamma}\|_{2,\boldsymbol{w}}^{2}=\sum_{e_{ij}\in E}w_{ij}(y_{ij}-\gamma_{ij}-\boldsymbol{\beta}^{T}\boldsymbol{\phi}_{i}+\boldsymbol{\beta}^{T}\boldsymbol{\phi}_{j})^{2},
\]
and $p_{\lambda}(\boldsymbol{\gamma})$ is  the sparsity constraint on $\boldsymbol{\gamma}$.  With this cost function, our Unified Robust Learning to Rank (URLR) framework identifies outliers globally by integrating all local pairwise comparisons together. \red{Note that in Eq (\ref{eq:HLasso_general}), the noise term $\boldsymbol{\epsilon}$ has been removed because the discrepancy is mainly caused by outliers due to their larger magnitude.}

Ideally the sparsity enhancing penalty term $p_{\lambda}(\boldsymbol{\gamma})$ should be a $l_{0}$ regularisation term. However, for a tractable solution, a $l_{1}$ 
regularisation term is used: $p_{\lambda}(\boldsymbol{\gamma})=\lambda\|\boldsymbol{\gamma}\|_{1,\boldsymbol{w}}=\lambda\sum_{e_{ij}} w_{ij}|\gamma_{ij}|$, where $\lambda$ is a free parameter corresponding to the weight for the regularisation term. 
With this $l_{1}$ penalty term,
the cost function becomes convex: 
\begin{equation}
L(\boldsymbol{\beta},\boldsymbol{\gamma})=\frac{1}{2}\|\sqrt{W}(\boldsymbol{y}-\boldsymbol{\gamma})-X\boldsymbol{\beta}\|_{2}^{2}+\lambda\|\boldsymbol{\gamma}\|_{1,\boldsymbol{w}},\label{eq:lagrange}
\end{equation}
where $X=\sqrt{W}C\Phi$, $W=\mathrm{diag}(\boldsymbol{w})$ is the diagonal matrix of $\boldsymbol{w}$ and $\sqrt{W}=\mathrm{diag}(\sqrt{\boldsymbol{w}})$.

Setting $\frac{\partial L}{\partial\boldsymbol{\beta}}=0$, the problem of minimisation of the cost function in \eqref{eq:lagrange} can be decomposed into the following two
subproblems: 
\begin{enumerate}
\item Estimating the parameters $\boldsymbol{\beta}$ of the prediction function $f(x)$: 
\begin{equation}
\hat{\boldsymbol{\beta}}=(X^{T}X)^{\dagger}X^{T}\sqrt{W}(\boldsymbol{y}-\boldsymbol{\gamma}),\label{eq:beta_org}
\end{equation}
 Mathematically, the Moore-Penrose pseudo-inverse of $X^{T}X$ is
defined as $(X^{T}X)^{\dagger}=\underset{\mu\rightarrow0}{\lim}((X^{T}X)^{T}(X^{T}X)+\mu I)^{-1}(X^{T}X)^{T}$,
where $I$ is the identity matrix. The scalar variable $\mu$ is introduced to avoid numerical instability
 \cite{elements_SL}, and typically assumes a small value\footnote{In this work, $\mu$ is set to $0.001$.%
}. With the the introduction of $\mu$, Eq (\ref{eq:beta_org}) becomes: 
\begin{equation}
\hat{\boldsymbol{\beta}}=(X^{T}X+\mu I)^{-1}X^{T}\sqrt{W}(\boldsymbol{y}-\boldsymbol{\gamma}).\label{eq:beta_org-1}
\end{equation}
A standard solver for Eq (\ref{eq:beta_org-1}) has a $O(|E|d^{2})$
computational complexity, which is almost linear with respect to the
size of the graph $|E|$ if $d\ll n$. Faster algorithms based
on the Krylov iterative and algebraic multi-grid methods \cite{anil2010largescale}
can also be used.
\item Outlier detection:
\begin{eqnarray}
\hat{\boldsymbol{\gamma}}= & \textrm{arg}\min_{\boldsymbol{\gamma}}\frac{1}{2}\Vert(I-H)\sqrt{W}(\boldsymbol{y}-\boldsymbol{\gamma})\Vert_{2}^{2}+\lambda\|\boldsymbol{\gamma}\|_{1,\boldsymbol{w}}\label{eq:gamma}\\
= & \textrm{arg}\min_{\boldsymbol{\gamma}}\frac{1}{2}\Vert\tilde{\boldsymbol{y}}-\widetilde{X}\boldsymbol{\gamma})\Vert_{2}^{2}+\lambda\|\boldsymbol{\gamma}\|_{1,\boldsymbol{w}}\label{eq:compute_outlier}
\end{eqnarray}
where $H=X(X^{T}X)^{\dagger}X^{T}$ is the hat matrix, $\widetilde{X}=(I-H)\sqrt{W}$
and $\tilde{\boldsymbol{y}}=\tilde{X}\boldsymbol{y}$.  Eq \eqref{eq:gamma}
is obtained by plugging the solution $\hat{\boldsymbol{\beta}}$ back
into Eq \eqref{eq:lagrange}. 
\end{enumerate}


\vspace{-0.3cm}
\subsection{Outlier detection by regularisation path}
\label{sub:Outlier-detection-by}
From the formulations described above, it is clear that outlier detection by solving Eq (\ref{eq:compute_outlier}) is the key -- once the outliers are identified, the estimated $\hat{\boldsymbol{\gamma}}$ can be used to substitute $\boldsymbol{\gamma}$ in Eq (\ref{eq:beta_org}) and the estimation of the prediction function parameter $\boldsymbol{\beta}$ becomes straightforward. Now let us focus on solving  Eq (\ref{eq:compute_outlier}) for outlier detection. 

Note that solving  Eq (\ref{eq:compute_outlier}) is essentially a LASSO
(Least Absolute Shrinkage and Selection Operator) \cite{lasso} problem. For a LASSO problem, tuning the regularisation parameter $\lambda$  
is notoriously difficult \cite{Huber81,scaledlasso,sqrtlasso,penalizedlasso}. In particular, 
in our URLR framework, the $\lambda$ value directly decides the ratio
of outliers in the training set which is unknown. A number of methods
for determining $\lambda$ exist, but none is suitable for our formulation: 
\begin{enumerate}
\item Some heuristics rules on setting the value of $\lambda$ such as $\lambda=2.5\hat{\sigma}$ are popular
in existing robust ranking models such as the M-estimator~\cite{Huber81},
where $\hat{\sigma}$ is a Gaussian variance set manually  based on
human prior knowledge. However setting a constant $\lambda$ value
independent of dataset is far from optimal because the ratio of outliers
may vary for different crowdsourced datasets.
\item Cross validation is also not applicable here because each edge $e_{ij}$
is associated with a $\gamma_{ij}$ variable and any
held-out edge $e_{ij}$ also has an associated unknown variable $\gamma_{ij}$.
As a result, cross validation can only optimise part of the sparse
variables while leaving those for the held-out validation set undetermined.
\item Data adaptive techniques such as Scaled LASSO \cite{scaledlasso} and Square-Root LASSO \cite{sqrtlasso} typically generate over-estimates on the support set of outliers. Moreover, they rely on the homogeneous Gaussian noise assumption which is often not valid in practice. 
\item The other alternatives e.g.~Akaike information criterion (AIC) and
Bayesian information criterion (BIC) are often unstable in outlier detection LASSO problems \cite{penalizedlasso}\footnote{\red{We  found  empirically that the model automatically selected by BIC or AIC failed to detect any meaningful outliers in our experiments. For details of the experiments and a discussion on the issue of determining the outlier ratio, please visit the project webpage at \url{http://www.eecs.qmul.ac.uk/~yf300/ranking/index.html}}}. 
\end{enumerate}

This inspires us to sequentially consider all available solutions
for all sparse variables along the Regularisation Path (RP) by gradually
decreasing the value of the regularisation parameter $\lambda$ from
$\infty$ to $0$. Specifically, based on the piecewise-linearity
property of LASSO, a regularisation path can be efficiently
computed by the R-package ``glmnet" \cite{glmnet}\footnote{\url{http://cran.r-project.org/web/packages/glmnet/glmnet.pdf}}. When $\lambda=\infty$,
the regularisation parameter will strongly penalise outlier detection:
if any annotation is taken as an outlier, it will greatly\textcolor{red}{{}
}increase the value of the cost function in Eq (\ref{eq:compute_outlier}).
When $\lambda$ is changed from $\infty$ to $0$, LASSO%
\footnote{For a thorough discussion from a statistical perspective, please see
\cite{FanLi01,incidential_par,lars,penalizedlasso}.%
} will first select the variable subset accounting for the highest
deviations to the observations $\widetilde{X}$ in
Eq (\ref{eq:compute_outlier}). These high deviations should be assigned
higher priority to represent the nonzero elements%
\footnote{This is related with LASSO for covariate selection in a graph. Please
see \cite{nico2006statistics} for more details.%
} of $\boldsymbol{\gamma}$ of Eq (\ref{eq:regression_basic_outlier}),
because $\boldsymbol{\gamma}$ compensates the discrepancy between
annotation and prediction. Based on this idea, we can order the edge
set $E$ according to which nonzero $\gamma_{ij}$
appears first when $\lambda$ is decreased from $\infty$ to $0$.
In other words, if an edge $e_{ij}$ whose associated outlier variable $\boldsymbol{\gamma}_{ij}$ becomes nonzero
at a larger $\lambda$ value, it has a higher probability to
be an outlier. Following this order, we identify the top $p\%$ edge
set $\Lambda_{p}$ as the annotation outliers. And its complementary
set $\Lambda_{1-p}=E\setminus\Lambda_{p}$ are the inliers. Therefore,
the outcome of estimating $\boldsymbol{\gamma}$ using Eq (\ref{eq:compute_outlier})
is a binary outlier indicator vector $\boldsymbol{f}=\left[f_{e_{ij}}\right]$:
\[
f_{e_{ij}}=\begin{cases}
\begin{array}{cc}
1 & e_{ij}\in\Lambda_{1-p}\\
0 & e_{ij}\in\Lambda_{p}
\end{array}\end{cases}
\]
where each element $f_{e_{ij}}$ indicates whether the
corresponding edge $e_{ij}$ is an outlier or not. 

Now with the outlier indicator vector $\boldsymbol{f}$ estimated using regularisation path, instead of  estimating $\boldsymbol{\beta}$ by substituting $\boldsymbol{\gamma}$ in Eq (\ref{eq:beta_org}) with  an estimated $\hat{\boldsymbol{\gamma}}$,
$\boldsymbol{\beta}$ can be computed as 
\begin{equation}
\hat{\boldsymbol{\beta}}=(X^{T}FX+\mu I)^{-1}X^{T}\sqrt{W}F\boldsymbol{y}\label{eq:beta_RP}
\end{equation}
where $F=\mathrm{diag}(\boldsymbol{f})$, that is, we use $\boldsymbol{f}$ to `clean up' $\boldsymbol{y}$ before estimating $\boldsymbol{\beta}$. 

The pseudo-code of learning
our URLR model is summarised  in Algorithm \ref{alg:Hodgerank-Algorithms.}.

\begin{algorithm}[t]
\textbf{Input}: A training dataset consisting of the feature matrix $\Phi$ and the pairwise annotation graph
$G$, and an outlier pruning rate $p\%$.

\textbf{Output}: Detected outliers $\boldsymbol{f}$ and prediction
model parameter $\boldsymbol{\beta}$. 
\begin{enumerate}
\item Solve Eq (\ref{eq:compute_outlier})  using Regularisation Path;
\item Take the top $p\%$ pairs as outliers to obtain the outlier indicator vector $\boldsymbol{f}$; 
\item Compute $\boldsymbol{\beta}$ using Eq (\ref{eq:beta_RP}). 
\end{enumerate}
\protect\caption{\label{alg:Hodgerank-Algorithms.}Learning a unified robust learning
to rank (URLR) model for SVP prediction}
\end{algorithm}

\subsection{Discussions}
\subsubsection{Advantage over majority voting}

\begin{figure}[h]
\begin{centering}
\includegraphics[scale=0.6]{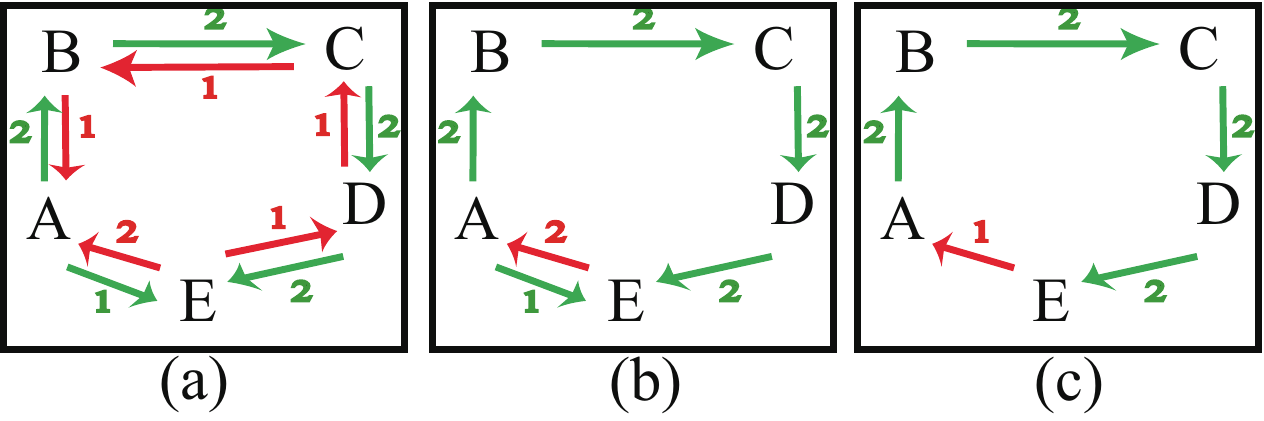}  
\par\end{centering}
\protect\caption{\label{fig:Two-example-to} Better outlier detection can be achieved
using our URLR framework than majority voting. Green arrows/edges indicate
correct annotations, while red arrows are outliers. The numbers indicate the number of votes received by each edge.}
\end{figure}

The proposed URLR framework identifies  outliers globally by integrating all
local pairwise comparisons together, in contrast to the local aggregation
based majority voting. 
Figure \ref{fig:Two-example-to}(a) illustrates why our
URLR framework is advantageous over the local majority voting method
for outlier detection. Assume there are five images $A-E$ with five
pairs of them compared three times each, and the correct global ranking order of these
5 images in terms of a specific SVP is $A<B<C<D<E$. Figure \ref{fig:Two-example-to}(a)
shows that among the five compared pairs, majority voting can successfully
identify four outlier cases: $A>B$, $B>C$, $C>D$, and $D>E$, but
not the fifth one $E<A$. However when considered globally, it is
clear that $E<A$ is an outlier because if we have $A<B<C<D<E$, we
can deduce $A<E$. Our formulation can detect this tricky outlier.
More specifically, if the estimated $\boldsymbol{\beta}$ makes $\boldsymbol{\beta}^{T}\boldsymbol{\phi}_{A}-\boldsymbol{\beta}^{T}\boldsymbol{\phi}_{E}>0$,
it has a small local inconsistency cost for that minority vote edge
$A\rightarrow E$. However, such $\boldsymbol{\beta}$ value will be `propagated'
to other images by using the voting edges $B\rightarrow A$, $C\rightarrow B$,
$D\rightarrow C$, and $E\rightarrow D$, which are accumulated into
a much bigger global inconsistency with the annotation. This enables our
model to detect $E\rightarrow A$ as an outlier, contrary to the majority
voting decision. In particular, the majority voting will introduce
a loop comparison $A<B<C<D<E<A$ which is the well-known Condorcet's
paradox \cite{condorcet,Hodgerank}. 

We further give two more extreme cases
in Figures \ref{fig:Two-example-to}(b) and (c). Due to the  Condorcet's
paradox, in Figure \ref{fig:Two-example-to}(b) the estimated $\boldsymbol{\beta}$
from majority voting, which removes $A\rightarrow E$,  is even worse than that from all annotation pairs
which at least save the correct annotation $A\rightarrow E$. Furthermore,
\red{Figure \ref{fig:Two-example-to}(c) shows that when each pair only
receives votes in one direction, majority voting will cease to
work altogether, but our URLR can still detect outliers by examining
the global cost.} This example thus highlights the capability of URLR in coping with extremely sparse pairwise comparison labels. In our experiments (see Section \ref{sec:Experiments}), the advantage of URLR over majority is validated on various SVP prediction problems. 

\subsubsection{Advantage over robust statistical ranking \label{sub:Connections-to-Huber-Lasso}}
\label{sec:over robust ranking}

Our framework is closely related to Huber's theory of robust regression \cite{Huber81}, which has been used for robust statistical ranking \cite{yuan13acmmm}. In contrast to learning to rank, robust statistical ranking is only concerned with ranking a set of training instances by integrating their (noisy) pairwise rankings. No low-level feature representation of the instances is used as robust ranking does not aim to learn a ranking prediction function that can be applied to unseen test data.  To see the connection between URLR with robust ranking, consider the Huber M-estimator
\cite{Huber81} which aims to estimate the optimal global ranking for a set of training instances by minimising the following cost function: 
\begin{equation}
\min_{\boldsymbol{\theta}}\sum_{i,j}w_{ij}\rho_{\lambda}((\theta_{i}-\theta_{j})-y_{ij})\label{eq:huberLoss}
\end{equation}
where $\boldsymbol{\theta} = [\theta_{i}]\in\mathbb{R}^{\mid E\mid} $  is the ranking score vector storing the global ranking score of each training instance
$i$. The Huber's loss function $\rho_{\lambda}(x)$ is defined as
\begin{equation}
\rho_{\lambda}(x)=\left\{ {\displaystyle \begin{array}{ll}
x^{2}/2, & \textrm{if \ensuremath{|x|\leq\lambda}}\\
\lambda|x|-\lambda^{2}/2, & \textrm{if \ensuremath{|x|>\lambda}.}
\end{array}}\right.
\end{equation}
Using this loss function, when $|(\theta_{i}-\theta_{j})-y_{ij}|<\lambda$,
the comparison is taken as a ``good'' one and penalised by an $l_{2}-$loss
for Gaussian noise. Otherwise, it is regarded as a sparse outlier
and penalised by an $l_{1}-$loss. 
It can be shown \cite{yuan13acmmm} that
 robust ranking with Huber's loss is equivalent to a LASSO problem,
which can been applied to joint robust ranking and outlier detection \cite{penalizedlasso}. Specifically, the global ranking of the training instances and the outliers in the pairwise rankings can be estimated as
\begin{eqnarray}
\left\{ \hat{\boldsymbol{\theta}},\hat{\boldsymbol{\gamma}}\right\} =\min_{\boldsymbol{\theta},\boldsymbol{\gamma}}\frac{1}{2}\|y-C\boldsymbol{\theta}-\boldsymbol{\gamma}\|_{2,\boldsymbol{w}}^{2}+\lambda\parallel\boldsymbol{\gamma}\parallel_{1,\boldsymbol{w}}\label{eq:HLasso-1}\\
=\mathrm{\min_{\boldsymbol{\theta},\boldsymbol{\gamma}}}\underset{e_{ij}\in E}{\sum}w_{ij}\left[\parallel\frac{1}{2}(y_{ij}-\gamma_{ij}-(\theta_{i}-\theta_{j})\parallel^{2}+\lambda|\gamma_{ij}|\right]\label{eq:HLasso}
\end{eqnarray}
The optimisation problem (\ref{eq:HLasso-1}) is designed for solving the
robust ranking problem with Huber's loss function, hence called Huber-LASSO
\cite{yuan13acmmm}.

Our URLR can be considered as a generalisation of the Huber-LASSO based robust ranking problem above. Comparing Eq (\ref{eq:HLasso-1}) with  Eq (\ref{eq:HLasso_general}), it can be seen that the main difference between URLR and  conventional robust ranking is that in URLR the cost function has the low-level feature matrix $\Phi$ computed from the training instances, and the prediction function parameter $\boldsymbol{\beta}$, such that $\theta=\Phi \boldsymbol{\beta}$. This is because the  objective of URLR is to predict SVP for unseen test data. However, URLR and robust ranking do share one thing in common -- the ability to detect outliers in the training data based on a Huber-LASSO formulation. This means that,  as opposed to our unified framework with feature $\Phi$,  one could design a two-step approach for learning to rank by first identifying and removing outliers using Eq (\ref{eq:HLasso-1}), followed by introducing the low-level feature matrix $\Phi$ and  prediction model parameter $\boldsymbol{\beta}$ and estimating $\boldsymbol{\beta}$ using Eq (\ref{eq:beta_RP}). We call this approach Huber-LASSO-FL based learning to rank which differs from URLR mainly in the way outliers are detected without considering low level features.

Next we show that there is a critical theoretical advantage of URLR over conventional Huber-LASSO in detecting outliers from the training instances. This is due to the difference in the projection space for estimating $\mathbb{\boldsymbol{\gamma}}$ which is denoted as $\Gamma$. To explain this point, we decompose $X$ in Eq (\ref{eq:compute_outlier})
by Singular Value Decomposition (SVD), 
\begin{equation}
X=\U\Sigma \mathcal{V}^{T}\label{eq:+}
\end{equation}
where $\U=[\U_{1},\U_{2}]$ with $\U_{1}$ being an orthogonal basis of the
column space of $X$ and $\U_2$ an orthogonal basis of its complement. 
Therefore, due to the orthogonality $\U^{T}\U=I$ and $\U_{2}^T X=0$, we
can simplify Eq (\ref{eq:compute_outlier}) into 
\begin{eqnarray}
\hat{\boldsymbol{\gamma}} & = & \textrm{arg}\min_{\boldsymbol{\gamma}}\Vert \U_{2}^{T}\boldsymbol{y}-\U_{2}^{T}\boldsymbol{\gamma}\Vert_{2,\boldsymbol{w}}^{2}+\lambda\Vert\boldsymbol{\gamma}\Vert_{1,\boldsymbol{w}}.\label{eq:outlier-1}
\end{eqnarray}

The SVD orthogonally projects $\boldsymbol{y}$ onto
the column space of $X$ and its complement, while $\U_{1}$ is an
orthogonal basis of the column space $X$ and $\U_{2}$ is the orthogonal
basis of its complement $\Gamma$ (i.e.~the kernel
space of $X^T$). With the SVD, we can now compute the outliers $\hat{\boldsymbol{\gamma}}$
by solving  Eq (\ref{eq:outlier-1}) which again is a LASSO problem \cite{elements_SL}, where outliers provide sparse approximations of projection $\U_2^T \boldsymbol{y}$. We can thus compare dimensions of the projection spaces of URLR and Huber-LASSO-FL:

\begin{itemize}
\item Robust ranking based on the featureless Huber-LASSO-FL\footnote{We assume that the graph is connected, that is,  $\left|E\right|\geq\left|V\right|-1$; we thus have $\mathrm{rank}(C)=\left|V\right|-1$.%
}: to see the dimension of the projection space $\Gamma$, i.e. the space of cyclic rankings \cite{Hodgerank,yuan13acmmm}, we can perform a similar SVD operation and rewrite Eq (\ref{eq:HLasso-1}) in the same form as Eq (\ref{eq:outlier-1}), but this time we have $X=\sqrt{W}C$, $\U_{1}\in\mathbb{R}^{\left|E\right|\times(\mid V\mid-1)}$
and $\U_{2}\in\mathbb{R}^{\left|E\right|\times(\left|E\right|-\left|V\right|+1)}$.
So the dimension of $\Gamma$ for Huber-LASSO-FL is $\mathrm{dim}(\Gamma)=\left|E\right|-\left|V\right|+1$.
\item URLR: in contrast we have $X=\sqrt{W}C\Phi$, $\U_{1}\in\mathbb{R}^{\mid E\mid\times d}$
and $\U_{2}\in\mathbb{R}^{\left|E\right|\times(\left|E\right|-d)}$.
So the dimension of $\Gamma$ for URLR is $\mathrm{dim}(\Gamma)=\left|E\right|-d$.
\end{itemize}
From the above analysis we can see that given a very sparse graph with $|E|\sim|V|$, the projection space $\Gamma$ for Huber-LASSO-FL will have a dimension ($\left|E\right|-\left|V\right|+1$) too small to be effective 
for detecting outliers. In contrast, by exploiting a low dimensional ($d\ll |V|$)
feature representation of the original node space, URLR can enlarge the
projection space to that of dimension $\left|E\right|-d$. 
Our URLR is thus able to enlarges its outlier detection projection space $\Gamma$. 
As a result our URLR can better identify outliers, especially for sparse pairwise annotation graphs.
In general, this advantage exists when the feature dimension $d$ is smaller than the number of training instance $\left|V\right| =N$, and 
the smaller the value of $d$, the bigger the advantage over Huber-LASSO. In practice,  given a large training set we typically have 
$d\ll\left|V\right|$. On the other hand, when the number of instances is small, and each instance is represented by a high-dimensional feature vector, we can always reduce the feature dimension using techniques such as  PCA to make sure that $d\ll\left|V\right|$. This theoretical advantage of URLR over conventional Huber-LASSO in outlier detection is validated experimentally in Section \ref{sec:Experiments}. 

\vspace{-0.3cm}
\subsubsection{Regularisation on $\boldsymbol{\beta}$}

It is worth mentioning that in the cost function of URLR (Eq (\ref{eq:HLasso_general})), there are two sets of variables to be estimated, $\boldsymbol{\gamma}$ and $\boldsymbol{\beta}$,  but only one $l_1$ regularisation term on $\boldsymbol{\gamma}$ to enforce sparsity. When the dimensionality of $\boldsymbol{\beta}$ (i.e.~$d$) is high, one would expect to see a $l_2$ regularisation term on $\boldsymbol{\beta}$ (e.g.~ridge regression) due to \red{the fact that the coefficients of highly correlated low-level features  can be poorly estimated and exhibit high variance  without imposing a proper size constraint on the coefficients~\cite{elements_SL}. 
The reason we do not include such a regularisation term is because, as mentioned above, using URLR we need to make sure the low-level feature space dimensionality $d$ is low, which means that the dimensionality of  $\boldsymbol{\beta}$ is also low, making the regularisation term $\boldsymbol{\beta}$ redundant. This leads to the applicability of much simpler solvers and we show empirically in the next section that satisfactory results can be obtained with this simplification.  }

\begin{table*}[t!]
\begin{centering}
\begin{tabular}{c||c|c|c|c}
\hline 
Dataset  & No. pairs  & No. img/video  & Feature Dim.  & No. classes\tabularnewline
\hline 
\hline 
Image Interestingness \cite{Isola2011cvpr}  & $16000$ & 2222  & 932 (150)  & 1\tabularnewline
\hline 
Video Interestingness \cite{yugangVideoInteresting2013}  & $60000$ & 420  & 1000 (60)  & 14\tabularnewline
\hline 
PubFig \cite{kumar2009,whittlesearch}  & $2616$ & 772  & 557 (100)  & 8\tabularnewline
\hline 
Scene \cite{scene_OSR,whittlesearch}  & $1378$ & 2688  & 512 (100)  & 8\tabularnewline
\hline 
FG-Net Face Age Dataset \cite{fu2010ageSurvey}  & --  & 1002  & 55  & --\tabularnewline
\hline 
\end{tabular}
\par\end{centering}
\protect\protect\caption{\label{tab:Dataset-summary.}Dataset summary.~We use the original
features to learn the ranking model (Eq (\ref{eq:beta_RP})) and
reduce the feature dimension (values in brackets) using Kernel PCA \cite{Mika99kernelpca} to improve
outlier detection (Eq (\ref{eq:compute_outlier})) by enlarging the
projection space of $\boldsymbol{\gamma}$. }
\end{table*}

\vspace{-0.3cm}
\section{Experiments \label{sec:Experiments}}

Experiments were carried out on five benchmark datasets (see Table \ref{tab:Dataset-summary.}) which fall into three categories: (1)  experiments on estimating subjective visual properties (SVPs) that are useful on their own including image (Section \ref{sec:image interestingness}) and video interestingness (Section \ref{sec:video interestingness}), (2)  experiments on estimating SVPs as relative attributes for visual recognition (Section \ref{sec:relative attributes}), and (3) experiments on human age estimation from face images (Section \ref{sec:age estimation}). The third set of experiments can be considered as synthetic experiments -- human age is not a subjective visual property although it is ambiguous and poses a problem even for humans \cite{fu2010ageSurvey}. However, as ground truth is available, this set of experiments are designed to gain insights into how different SVP prediction models work.

\begin{figure*}
\begin{centering}
\includegraphics[scale=0.4]{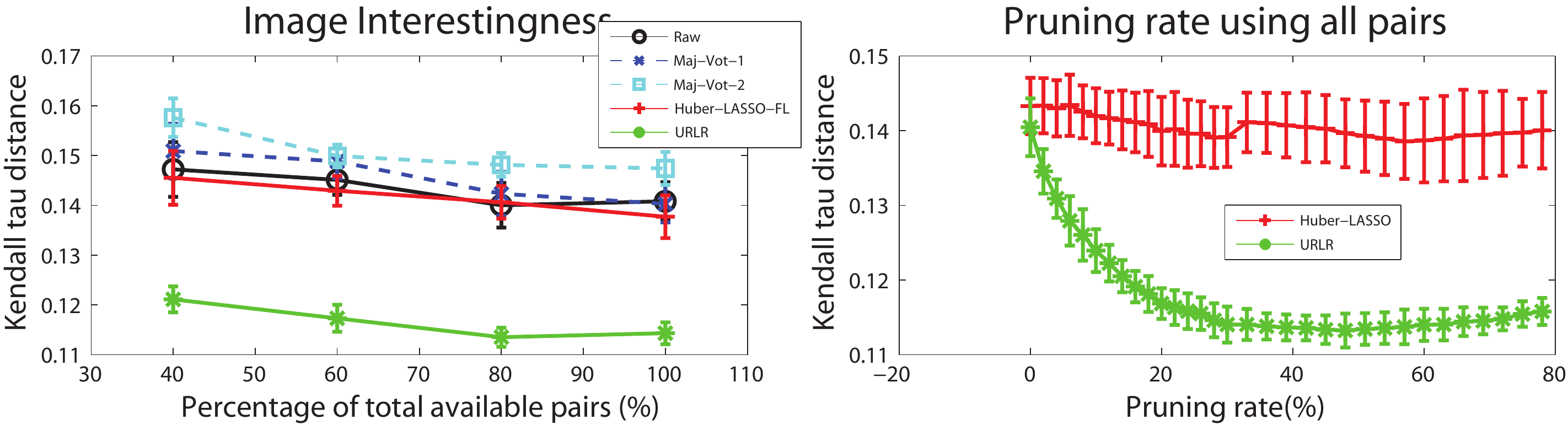}  
\par\end{centering}
\vspace{-0.3cm}
\protect\protect\caption{\label{fig:res_img_interestingness} Image interestingness prediction
comparative evaluation. Smaller Kendall tau distance means better performance. The mean and standard deviation of each method over 10 trials  are shown in the plots. }
\end{figure*}

\noindent 
\begin{figure*}
\begin{centering}
\includegraphics[scale=0.4]{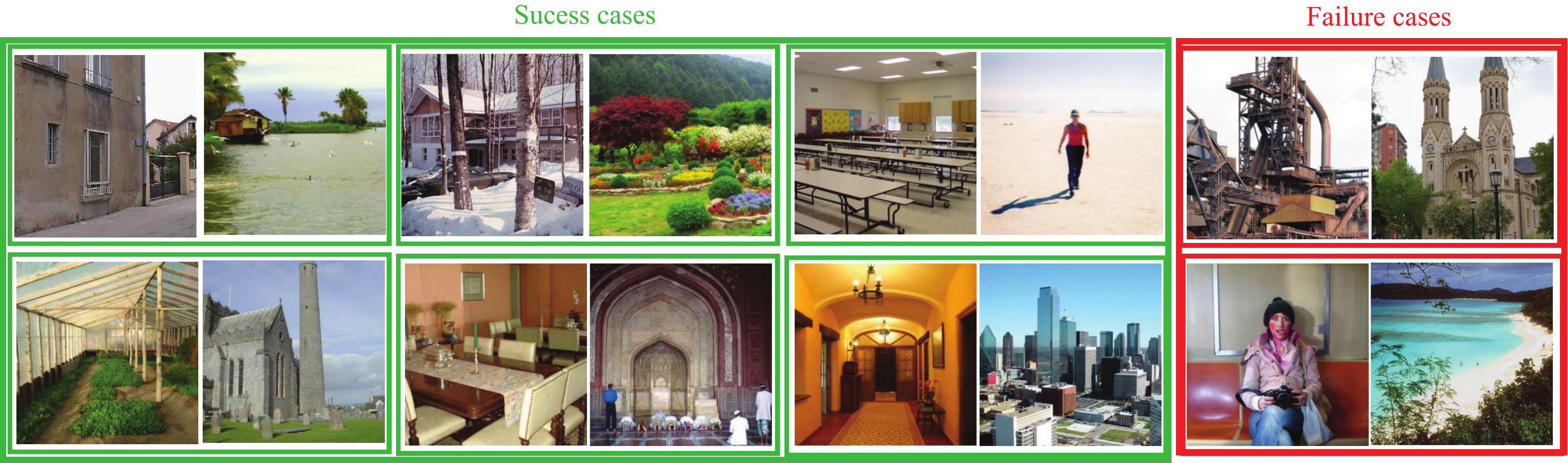}   
\par\end{centering}
\protect\protect\caption{\label{fig:Quantative-results-of} \red{Qualitative examples of outliers detected 
by URLR}. In each box, there are two images. The left image was annotated as more interesting than
the right. Success cases (green boxes) show true positive outliers
detected by URLR (i.e.~ right images are more interesting according to the ground truth). Two failure cases are shown in red boxes (URLR thinks the images on the right are more interesting but the ground truth agrees with the annotation).}
\end{figure*}

\vspace{-0.3cm}
\subsection{Image interestingness prediction}
\label{sec:image interestingness}

\noindent \textbf{Datasets}\quad{}  The~image~interestingness~dataset was first introduced
in \cite{Isola2011cvpr} for studying memorability. It was later re-annotated
as an image interestingness dataset by \cite{imginterestingnessICCV2013}.
It consists of $2222$ images. \red{Each was represented as a $915$ dimensional attribute\footnote{We delete $8$ attribute features from the original feature vector in \cite{Isola2011cvpr,imginterestingnessICCV2013} such as ``attractive'' because they are highly correlated with image interestingness.}
feature vector~\cite{Isola2011cvpr,imginterestingnessICCV2013} such as central object, unusual scene and so on. }
$16000$
pairwise comparisons were collected by \cite{imginterestingnessICCV2013}
using AMT and used as annotation.  On average, each image is viewed and compared with 11.9 other images, resulting a total of 16000 pairwise labels\footnote{\red{On average, for each labelled pair,  around 80\% of the annotations agree with one ranking order and 20\% the other.}}.

\noindent \textbf{Settings}\quad{}  $1000$ images were randomly selected for training and the remaining $1222$
 for testing. All the experiments were repeated $10$ times with different random training/test splits
to reduce variance. The pruning rate $p$ was set to $20\%$. We also varied the number of annotated pairs used to  test how well each compared method copes with increasing annotation sparsity.

\noindent \textbf{Evaluation metrics}\quad{} For both image and video interestingness
prediction, Kendall tau rank distance was employed to measure the percentage of pairwise mismatches between the predicted ranking order for each pair of test data using their prediction/ranking function scores, and the ground truth
ranking provided by \cite{imginterestingnessICCV2013} and \cite{yugangVideoInteresting2013} respectively.
Larger Kendall tau rank distance means
lower quality of the ranking order predicted.

\noindent \textbf{Competitors}\quad{}We compare our method (URLR)
with four competitors.
\begin{enumerate}
\item \noindent \textbf{\emph{Maj-Vot-1}}~\cite{yugangVideoInteresting2013}:
this method uses majority voting for outlier pruning and rankSVM for
learning to rank. 
\item \noindent \textbf{\emph{Maj-Vot-2}}~\cite{imginterestingnessICCV2013}:
this method also first removes outliers by majority voting. After
that, the fraction of selections by the pairwise comparisons for each
data point is used as an absolute interestingness score and a regression
model is then learned for prediction. Note that \emph{Maj-Vot-2} was
only compared in the experiments on image and video interestingness prediction, since
only these two datasets have enough dense annotations for \emph{Maj-Vot-2}.
\item \noindent \textbf{\emph{Huber-LASSO-FL}}\textbf{:} robust
statistical ranking  that performs outlier detection using the conventional featureless Huber-LASSO as described in Section 
\ref{sub:Connections-to-Huber-Lasso}, followed by estimating $\boldsymbol{\beta}$ using Eq (\ref{eq:beta_RP}). 
\item \noindent \textbf{\textit{Raw}}\textbf{:}  our URLR model without
outlier detection, that is, all annotations are used to estimate $\boldsymbol{\beta}$.
\end{enumerate}

\noindent \textbf{Comparative results}\quad{} The interestingness prediction performance of the various  models are evaluated while varying the  amount of pairwise annotation used.  The results are shown
in Figure \ref{fig:res_img_interestingness}~(left). It shows clearly
that our \emph{URLR} significantly outperforms the four alternatives
for a wide range of annotation density. This validates the effectiveness
of our method. In particular, it can be observed that: (1) The improvement over \emph{Maj-Vot-1}
\cite{yugangVideoInteresting2013} and \emph{Maj-Vot-2} \cite{imginterestingnessICCV2013}
demonstrates the superior outlier detection ability of \emph{URLR} due to global rather than local outlier detection.
(2) \emph{URLR} is superior to \emph{Huber-LASSO-FL} because the joint outlier
detection and ranking prediction framework of \emph{URLR} enables
the enlargement of the projection space $\Gamma$ for $\boldsymbol{\gamma}$ (see Section \ref{sec:over robust ranking}),
resulting in better outlier detection performance. (3) The performance
of \emph{Maj-Vot-2} \cite{imginterestingnessICCV2013} is the worst
among all methods compared, particularly so given sparser annotation.
This is not surprising -- in order to get an reliable absolute interestingness
value, dozens or even hundreds of comparisons per image are required,
a condition not met by this dataset. (4) The performance of \emph{Huber-LASSO-FL} is also better
than \emph{Maj-Vot-1} and \emph{Maj-Vot-2} suggesting even a weaker
global outlier detection approach is better then the majority voting
based local one. (5) Interestingly even the baseline method \emph{Raw}
gives a comparable result to \emph{Maj-Vot-1} and \emph{Maj-Vot-2}
which suggests that just using all annotations without discrimination
in a global cost function (Eq (\ref{eq:lagrange})) is as effective
as majority voting\footnote{\red{One intuitive explanation for this is that given a pair of data with multiple contradictory votes, using Raw, both the correct and incorrect votes contribute to the learned model. In contrast, with Maj-Vot, one of them is eliminated, effectively amplifying the other's contribution in comparison to Raw. When the ratio of outliers gets higher, Maj-Vot will make more mistakes in eliminating the correct votes. As a result, its performance drops to that of Raw, and eventually falls below it. } }. 

Figure \ref{fig:res_img_interestingness}~(right) evaluates how the
performances of \emph{URLR} and \emph{Huber-LASSO-FL} are affected by
the pruning rate $p$. It can be seen that the performance of \emph{URLR}
is improving with an increasing pruning rate. This means that our
\emph{URLR} can keep on detecting true positive outliers. The gap
between \emph{URLR} and \emph{Huber-LASSO-FL} gets bigger when more comparisons
are pruned showing \emph{Huber-LASSO-FL} stops detecting outliers much
earlier on. \red{However, when the pruning rate is over 55\%, since most outliers have been removed, inliers start to be pruned, leading to poorer performance}. 

\noindent \textbf{Qualitative Results}\quad{}Some examples of outlier detection using \emph{URLR}
are shown in Figure \ref{fig:Quantative-results-of}. It can be seen that those in the green boxes are clearly outliers and are detected correctly by our URLR. The failure cases are interesting. For example, in the bottom case,  ground truth indicates that the woman sitting on a bench is more interesting than the nice beach image, whilst our URLR predicts otherwise. \red{The odd facial appearance} on that woman or the fact that she is holding a camera could be the reason why this image is considered to be more interesting than the otherwise more visually appealing beach image. \red{However, it is unlikely that the features used by URLR  are powerful enough to describe such fine appearance details. }

\vspace{-0.4cm}
\subsection{Video interestingness prediction}
\label{sec:video interestingness}

\noindent \textbf{Datasets}\quad{} The~video~interestingness~dataset is the YouTube interestingness
dataset introduced in \cite{yugangVideoInteresting2013}. It contains
$14$  categories of advertisement videos (e.g. `food' and `digital products'), each of which has $30$ videos.
$10\sim15$ annotators were asked to give complete interesting comparisons
for all the videos in each category. So the original annotations are
noisy but not sparse. We used bag-of-words of Scale Invariant Feature
Transform (SIFT) and Mel-Frequency Cepstral Coefficient (MFCC) as
the feature representation which were shown to be effective in \cite{yugangVideoInteresting2013}
for predicting video interestingness.

\noindent \textbf{Experimental settings}\quad{}Because comparing
videos across different categories is not very meaningful, we followed
the same settings as in \cite{yugangVideoInteresting2013} and only
compared the interestingness of videos within the same category. Specifically,
from each category we used $20$ videos  and their paired
comparisons for training and the remaining $10$ videos for testing.
The experiments were repeated for $10$ rounds and the averaged results
are reported. 

Since MFCC and SIFT are bag-of-words features, we employed $\chi^{2}$
kernel to compute and combine the features. To facilitate the computation,
the $\chi^{2}$ kernel is approximated by additive kernel of explicit
feature mapping \cite{additiveKernel}. To make the results of this
dataset more comparable to those in \cite{yugangVideoInteresting2013},
we used rankSVM model to replace Eq (\ref{eq:beta_RP}) as the ranking
model. As in the image interestingness experiments, we used Kendal tau rank distance as the evaluation metric,
while we  find that the same results can be obtained if the prediction
accuracy in \cite{yugangVideoInteresting2013} is used. The pruning
rate was again set to $20\%$.

\begin{figure*}[t]
\begin{centering}
\includegraphics[scale=0.4]{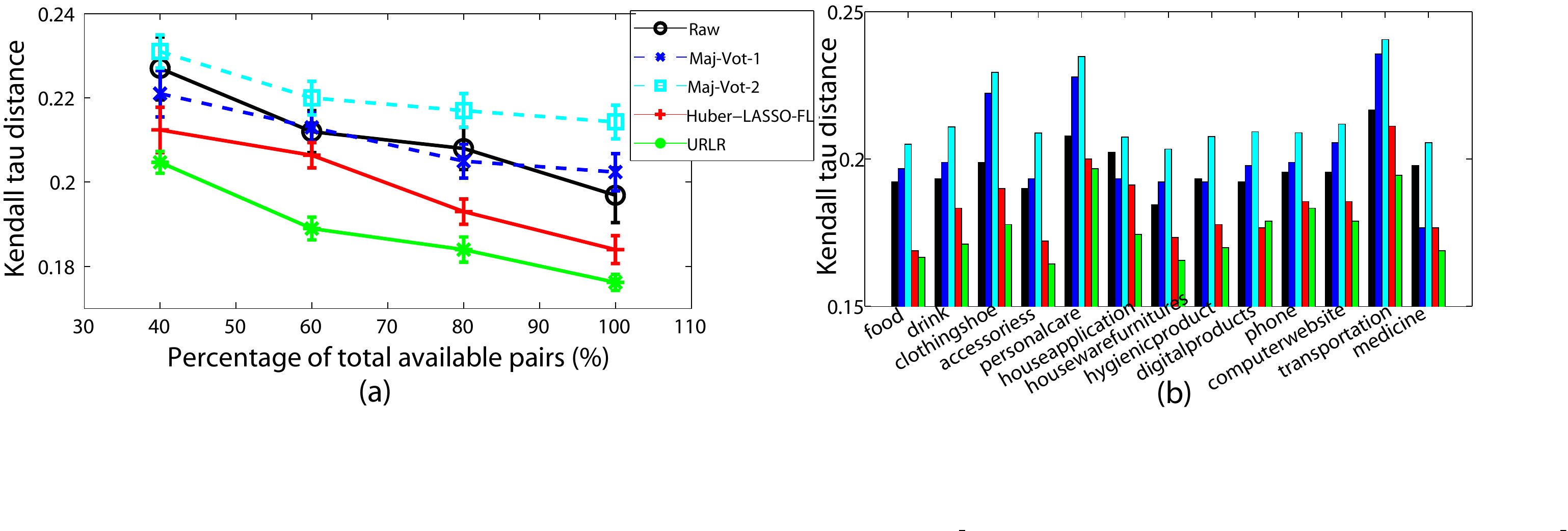}  
\par\end{centering}
\vspace{-0.3cm}
\protect\protect\caption{\label{fig:Youtube_video_interestingness}Video interestingness prediction
comparative evaluation. }
\end{figure*}

\noindent \textbf{Comparative Results}\quad{}
Figure \ref{fig:Youtube_video_interestingness}(a) compares the
 interestingness prediction methods given varying amounts of annotation, and Figure \ref{fig:Youtube_video_interestingness}(b)
shows the per category performance. The results
show that all the observations we had for the image interestingness
prediction experiment still hold here, and across all categories.
However in general the gaps between our \emph{URLR} and the alternatives
are smaller as this dataset is densely annotated. In particular the
performance of \emph{Huber-LASSO-FL} is much closer to our \emph{URLR}
now. This is because the advantage of \emph{URLR} over \emph{Huber-LASSO-FL}
is stronger when $|E|$ is close to $|V|$. In this experiment,
$|E|$ (1000s) is much greater than $|V|$ (20) and the advantage of enlarging the projection space $\Gamma$
for $\boldsymbol{\gamma}$ (see Section \ref{sec:over robust ranking}) diminishes. 

\begin{figure*}
\centering{}\includegraphics[scale=0.6]{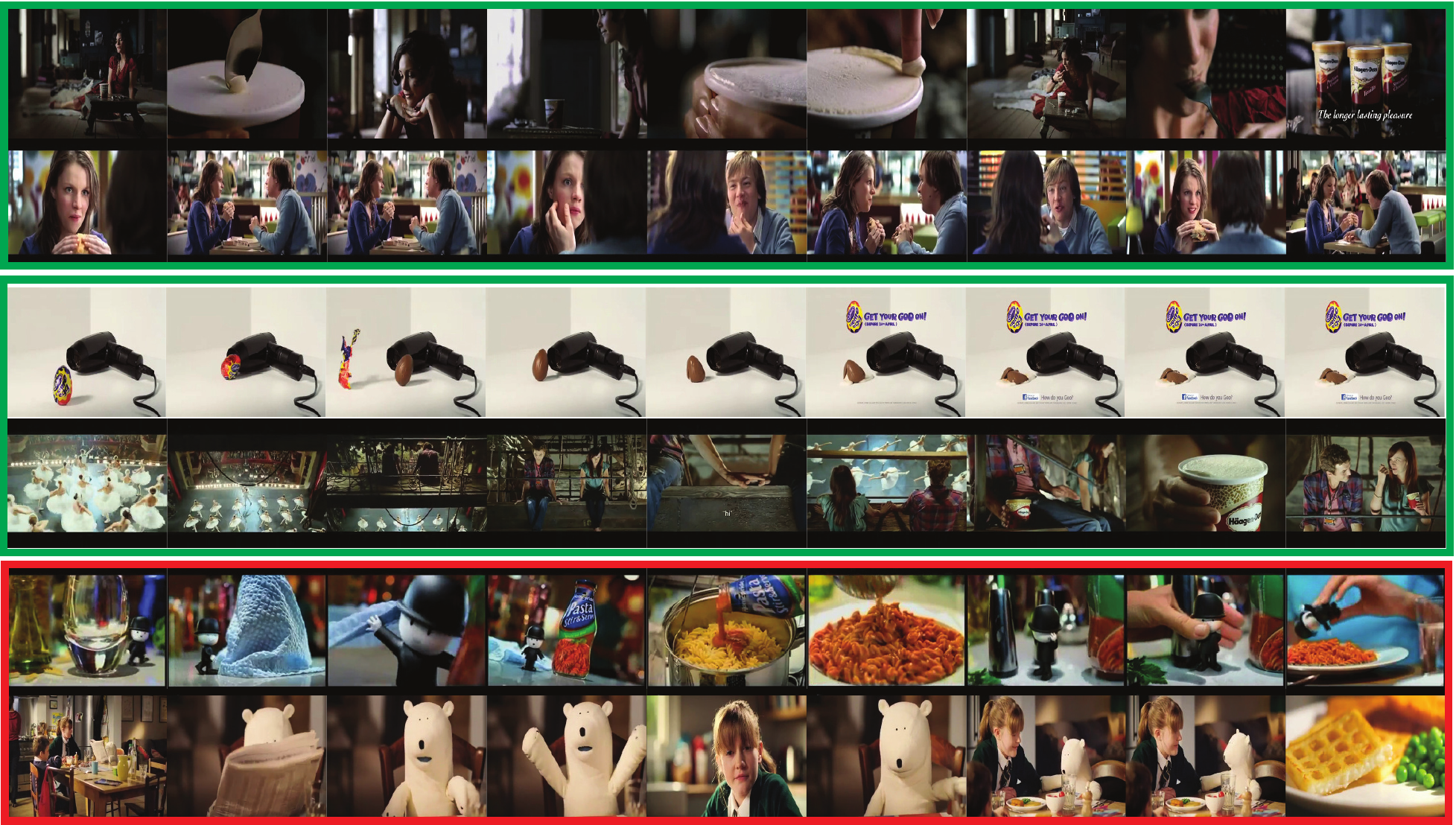} 
\protect\caption{\label{fig:Qualitative-results-on_videointerestingness}Qualitative
examples of video interestingness outlier detection. For each pair, the top video was annotated as more interesting than the bottom. Green boxes indicate the annotations are correctly detected as outliers
by our URLR and red box indicates a failure case (false positive). All 6 videos are from the `food' category.}
\end{figure*}

\noindent \textbf{Qualitative Results}\quad{}Some outlier detection examples are
shown in Figure \ref{fig:Qualitative-results-on_videointerestingness}.
In the two successful detection examples, the bottom videos are clearly more interesting than the top ones, because they (1) have a plot, sometimes with a twist, and (2) are accompanied by popular songs in the background and/or conversations. Note that in both cases,  majority voting would consider them inliners. The failure case is a hard one: both videos have cartoon characters, some plot, some conversation, and similar music in the background. This thus corresponds to a truly ambiguous case which can go either way. 

\vspace{-0.3cm}
\subsection{Relative attributes prediction}
\label{sec:relative attributes}

\begin{figure*}[t]
\centering{}\includegraphics[scale=0.5]{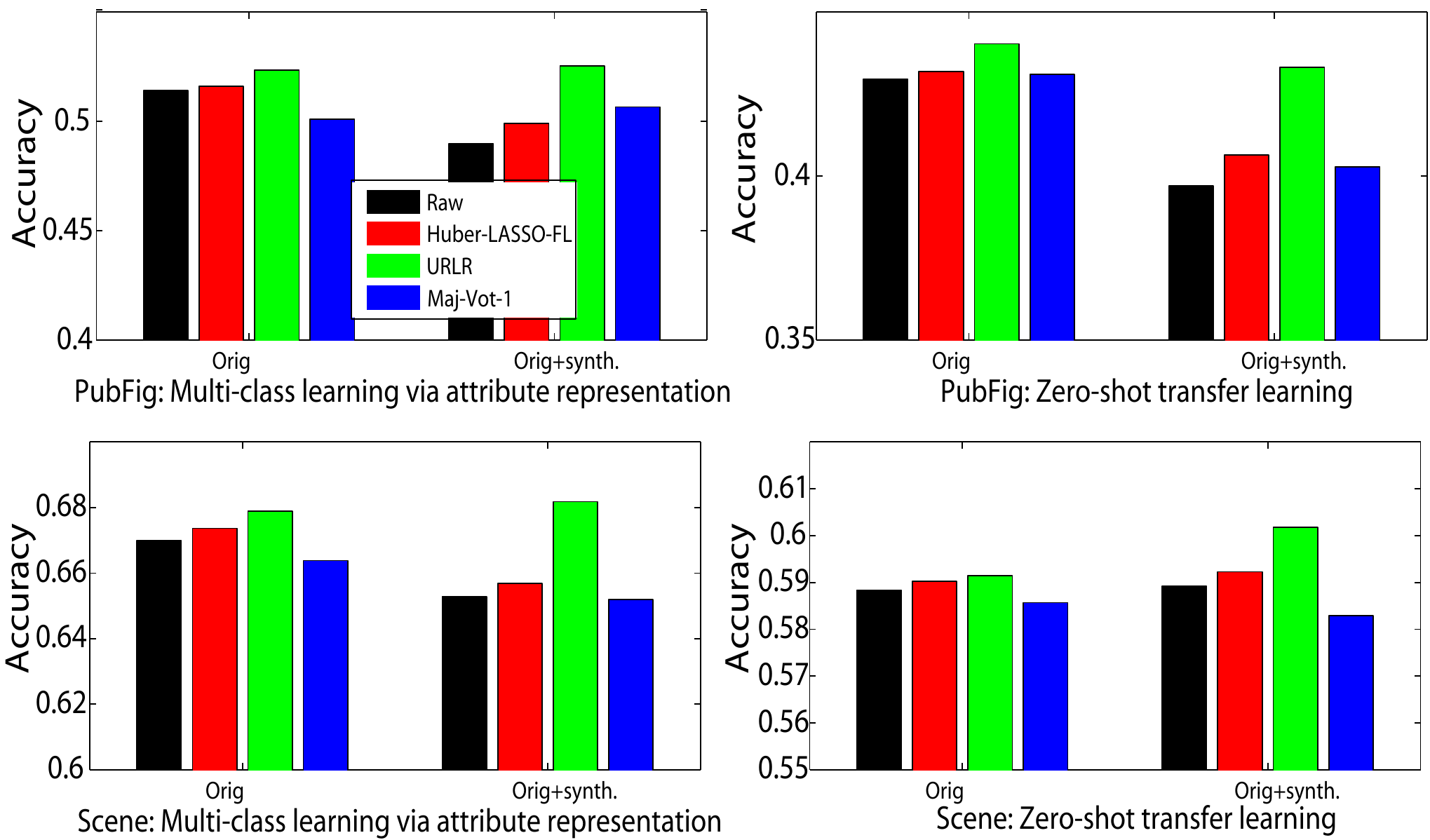}    
\protect\protect\caption{\label{fig:img1ZSC-1} Relative attribute performance evaluated indirectly
as image classification rate (chance = 0.125). }
\end{figure*}

\noindent \textbf{Datasets}\quad{} The \emph{PubFig}~\cite{kumar2009}~and~Scene~\cite{scene_OSR}~datasets
are two relative attribute datasets.  PubFig contains $772$ images from $8$ people and $11$ attributes
(\textquoteleft{}smiling\textquoteright{}, \textquoteleft{}round face\textquoteright{},
etc.). Scene \cite{scene_OSR} consists of $2688$ images from $8$
categories and $6$ attributes (\textquoteleft{}openness\textquoteright{},
\textquoteleft{}natrual\textquoteright{} etc.).  Pairwise attribute annotation was collected by Amazon Mechanical
Turk  \cite{whittlesearch}. Each pair was labelled by $5$ workers and majority vote was used in  \cite{whittlesearch} to average the comparisons for each pair\footnote{Thanks to the authors of \cite{whittlesearch} we have all the 
the raw pairs data before majority voting.%
}.  A total of $241$ and $240$ training images for PubFig and Scene respectively were labelled (i.e.~compared with at least another image). The average number of compared pairs per attribute were $418$ and $426$ respectively, meaning most images were only compared with one or two other images. The annotations for both datasets were thus  extremely sparse. 
GIST and colour histogram features were
used for PubFig, and GIST alone for Scene. Each image also belongs
to a class (different celebrities or scene types). These  datasets were designed
for classification, with the predicted relative attribute scores used as image representation. 

\noindent \textbf{Experimental Settings}\quad{}We evaluated two different image
classification tasks: multi-class classification
where samples from all classes were available for training and zero-shot
transfer learning where one class was held out during training (a different
class was used in each trial with the result averaged). Our experiment
setting was similar to that in \cite{parikh2011relativeattrib}, except
that image-level, rather than class-level pairwise comparisons were
used. Two settings were used with different amounts of annotation noise: 
\begin{itemize}
\item \emph{Orig:} This was the original setting with the pairwise annotations
used as they were. 
\item \emph{Orig+synth}: By visual inspection, there were limited annotation
outliers in these datasets, perhaps because these relative attributes
are less subjective compared to interestingness. To simulate more
challenging situations, we added $150$ random comparisons for
each attribute, many of which would correspond to outliers. This will
lead to around $20\%$ extra outliers. 
\end{itemize}
\noindent The pruning rate was set to $7\%$ for the original datasets (\emph{Orig})
and $27\%$ for the dataset with additional outliers inserted for all
attributes of both datasets (\emph{Orig+synth}).

\noindent \textbf{Evaluation metrics}\quad{} For Scene and Pubfig 
datasets,  relative attributes were very sparsely collected and their
prediction performance is thus evaluated indirectly by image classification
accuracy with the predicted relative attributes as image representation. Note that for image classification there is ground truth and its accuracy is clearly dependent on the relative attribute prediction
accuracy. For both datasets, we employed the method in \cite{parikh2011relativeattrib} to compute the image classification accuracy.

\noindent \textbf{Comparative Results}\quad{}Without the ground truth
of relative attribute values, different models were evaluated indirectly
via image classification accuracy in Figure~\ref{fig:img1ZSC-1}. The
following observations can be made: (1) Our URLR always outperforms
\emph{Huber-LASSO-FL}, \emph{Maj-Vot-1} and \emph{Raw} for all experiment
settings. The improvement is more significant when the data contain
more errors (\emph{Orig+synth}). (2) The performance of other methods
is in general consistent to what we observed in the image and video
interestingness experiments: \emph{Huber-LASSO-FL} is better than \emph{Maj-Vot-1}
and Raw often gives better results than majority voting. (3) For PubFig, \emph{Maj-Vot-1} \cite{yugangVideoInteresting2013}
is better than \emph{Raw} given more outliers, but it is not the case
for Scene. This is probably because the annotators were more familiar
with the celebrity faces in PubFig and hence their attributes than those
in Scene. Consequently there should be more subjective/intentional
errors for Scene, causing majority voting to choose wrong local ranking
orders (e.g. some people are unsure how to compare the relative
values of the `diagonal plane' attribute for two images). These majority
voting + outlier cases can only be rectified by using a global approach
such as our \emph{URLR}, and \emph{Huber-LASSO-FL} to a lesser
extent.

\begin{figure*}[t]
\begin{centering}
\includegraphics[scale=1.3]{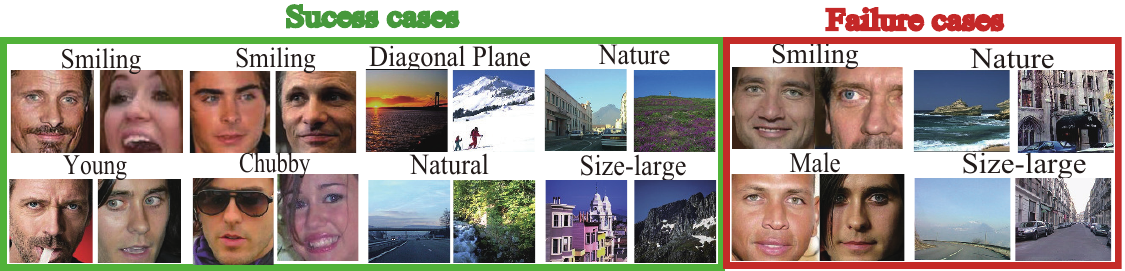} 
\par\end{centering}
\protect\protect\caption{\label{fig:Examples-of-outliers}Qualitative results on image relative
attribute prediction.}
\end{figure*}

\noindent \textbf{Qualitative Results}\quad{}Figure \ref{fig:Examples-of-outliers}
gives some examples of the pruned pairs for both datasets using URLR.
In the success cases, the left images were (incorrectly) annotated
to have more of the attribute than the right ones. However, they are
either wrong or too ambiguous to give consistent answers, and as such
are detrimental to learning to rank. A number of failure cases (false
positive pairs identified by \emph{URLR}) are also shown. Some of
them are caused by unique view point (e.g.~Hugh Laurie's mouth is
not visible, so it is hard to tell who smiles more; the building and
the street scene are too zoomed in compared to most other samples);
others are caused by the weak feature representation, e.g.~in the
`male' attribute example, the colour and GIST features are not discriminative
enough for judging which of the two men has more `male' attribute.

\noindent \textbf{Running Cost}\quad{} Our algorithm is very efficient with a unified framework where all outliers are pruned simultaneously and the ranking function estimation has a closed form solution. Using URLR on PubFig, it took only 1 minutes to prune 240 images with 10722 comparisons and learn the ranking function for  attribute prediction on a PC with four 3.3GHz CPU cores and 8GB memory.

\vspace{-0.3cm}
\subsection{Human age prediction from face images}
\label{sec:age estimation}

In this experiment, we consider age as a subjective visual property of a face. This is  partially true -- for many people, given a face image predicting the person's age can be subjective. The key difference between this and the other SVPs evaluated so far is that we do have the ground truth, i.e.~the person's age when the picture was taken. This enables us to perform in-depth evaluation of the 
significance of our URLR framework over the alternatives on various factors such as annotation sparsity,  and outlier ratio (we now know the exact ratio). Outlier detection accuracy can also now be measured directly.

\noindent \textbf{Dataset} \quad The  FG-NET image
age dataset\footnote{http://www.fgnet.rsunit.com/} was employed which contains $1002$ images of $82$
individuals labelled with ground truth ages ranging from  $0$ to $69$. The training set is composed
of the images of $41$ randomly selected people and the rest used as the test set.
All experiments were repeated $10$ times with different training/testing splits
to reduce variability. Each image was represented by a $55$ dimension
vector extracted by active appearance models (AAM) \cite{fu2010ageSurvey}.

\noindent \textbf{Crowdsourcing errors} We used the ground truth age
to generate the pairwise comparisons without any error.
Errors were then synthesised according to human error patterns estimated
by data collected by an online pilot study\footnote{\url{http://www.eecs.qmul.ac.uk/~yf300/survey4/}}:
\red{$4000$ pairwise image comparisons from $20$  willingly participating
``good'' workers were collected as \textit{unintentional errors.}  So we assume they are not contributing random or
malicious annotations. Thus the errors of these pairwise comparisons come from the natural data ambiguity.   
 The human unintentional age error pattern was built by fitting the
error rate against true age difference between collected pairs. As
expected, humans are more error-prone for smaller age difference.  Specifically, we fit quadratic polynomial function to model relation of age difference of two samples towards the chance of making an unintentional error.  
We then used this error pattern to generate unintentional errors.}  \emph{I}\textit{ntentional
errors} were introduced by `bad' workers who provided  random pairwise labels.
This was easily simulated by adding random comparisons. In practice,
human errors in crowdsourcing experiments can be a mixture of both
types. Thus two settings were considered: \textit{Unint.}: errors were
generated following the estimated human unintentional error model
resulting in around $10\%$ errors. \textit{Unint.+Int.}: random comparisons
were added on top of \textit{Unint.}, giving an error ratio of around
$25\%$, unless otherwise stated.  \red{ Since the ground-truth age of each face image is known to us, 
we can give an upper bound for all the compared methods by using ground-truth age of training data to generate a set of pairwise comparisons. This outlier-free dataset is then used to learn a kernel ridge regression with Gaussian kernel. This ground-truth data trained model is denoted as \emph{GT}.
}

\noindent \textbf{Quantitative results} Four experiments were conducted
using different settings to show the effectiveness of our URLR method quantitatively.

\begin{figure}
\begin{centering}
\includegraphics[scale=0.4]{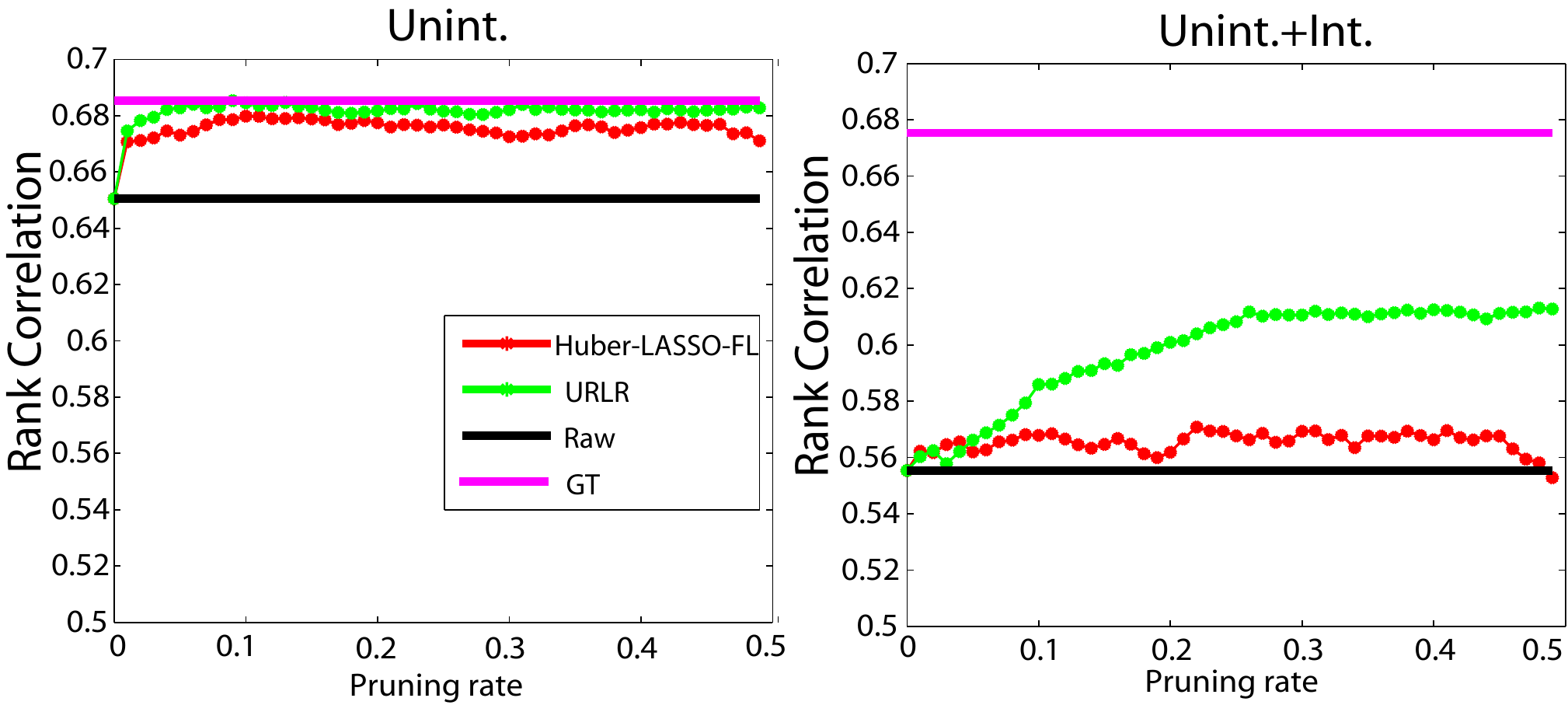}   
\par\end{centering}

\vspace{-0.3cm}
 \protect\protect\caption{\label{fig:rank_correlation_FGNET}Comparing \emph{URLR }and\emph{
Huber-LASSO-FL} on ranking prediction under two error settings. Note that the ranking prediction accuracy is measured using Kendall tau rank correlation which is very similar to Kendall tau distance (see \cite{rankcorrelation}). With rank correlation, the higher the value the better the performance.  }
\end{figure}

\noindent \emph{(1) URLR vs. Huber-LASSO-FL.} In this experiment, $300$ training
images and $600$ unique comparisons were randomly sampled from the training set. Figure \ref{fig:rank_correlation_FGNET}
shows that \emph{URLR} and \emph{Huber-LASSO-FL} improve over \emph{Raw} indicating
that outliers are effectively pruned using both global outlier detection methods. Both methods are robust to low
error rate~(Figure \ref{fig:rank_correlation_FGNET} Left: $10\%$
in Unint.) and are fairly close to GT, whilst the performance of \emph{URLR} is significantly better
than that of \emph{Huber-LASSO-FL} given high error ratio (Figure \ref{fig:rank_correlation_FGNET}
Right: $25\%$ in Unint.+Int.) because of the using low-level feature
representation to increase the dimension of projection space dimension for $\boldsymbol{\gamma}$ from $301$ for \emph{Huber-LASSO-FL} to
$546$ for \emph{URLR} (see Section \ref{sec:over robust ranking}). This result again validates our analysis that higher
$\mathrm{dim}(\Gamma)$ leads to better chance of identifying  outliers correctly. It is noted that  in Figure \ref{fig:rank_correlation_FGNET}(Right), given $25\%$ outliers, the result indeed peaks when $p$ is around 25; importantly, it stays flat when up to 50\% of the annotations are pruned. 

\begin{figure}
\begin{centering}
\includegraphics[scale=0.4]{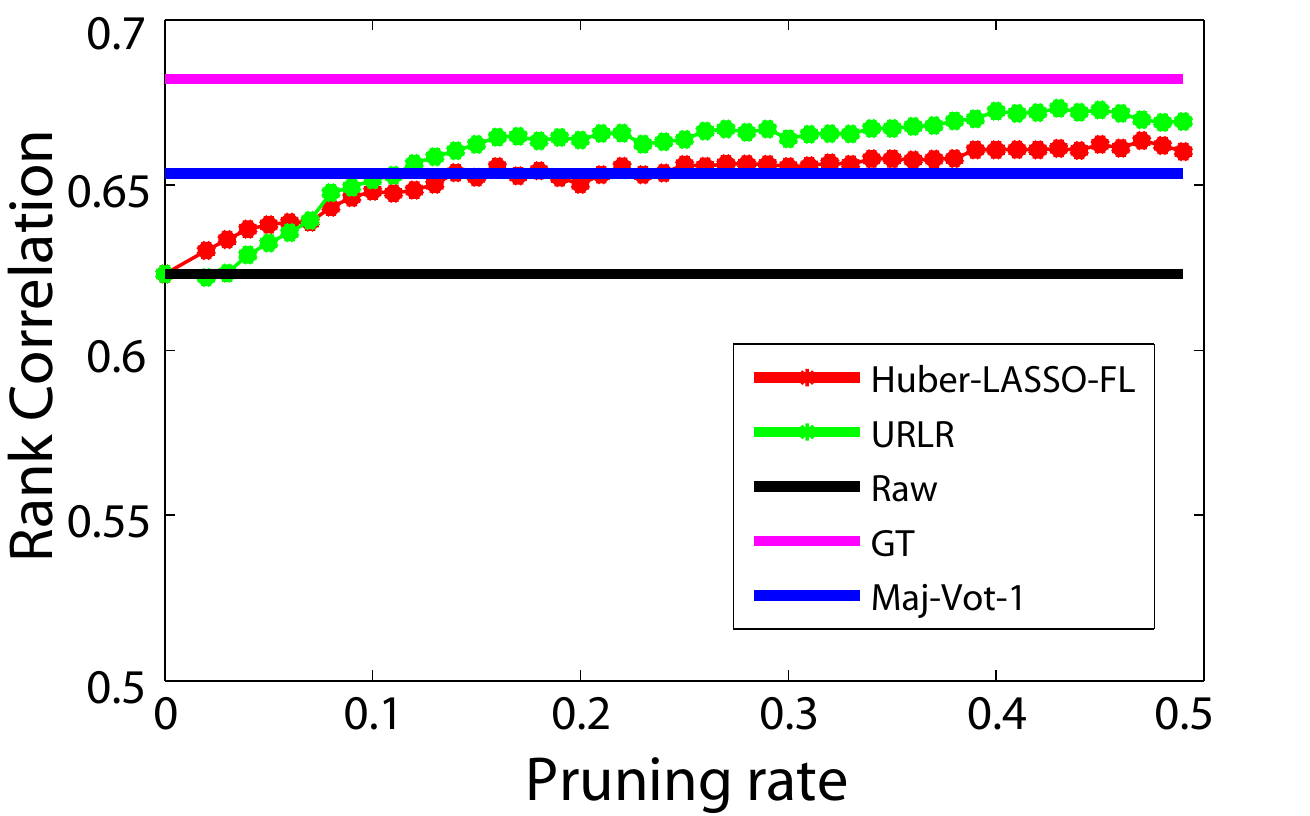}   
\par\end{centering}

\vspace{-0.3cm}
 \protect\protect\caption{\label{fig:Majority-voting.}Comparing \emph{URLR} and \emph{Huber-LASSO-FL}
against majority voting (5 comparisons per pair).}
\end{figure}

\noindent \emph{(2) Comparison with Maj-Vot-1.} Given the same data
but each pair compared by $5$ workers (instead of $1$) under the Unint.+Int.~error
condition, Figure \ref{fig:Majority-voting.} shows that \emph{Maj-Vot-1}
beats \emph{Raw}. This shows that for relative dense graph, majority
voting is still a good strategy of removing some outliers and improves
the prediction accuracy. However, \emph{URLR} outperforms \emph{Maj-Vot-1}
after the pruning rate passes $10\%$. This demonstrates that aggregating
all paired comparisons globally for outlier pruning is more effective
than aggregating them locally for each edge as done by majority voting.

\begin{figure}
\centering{}\includegraphics[scale=0.45]{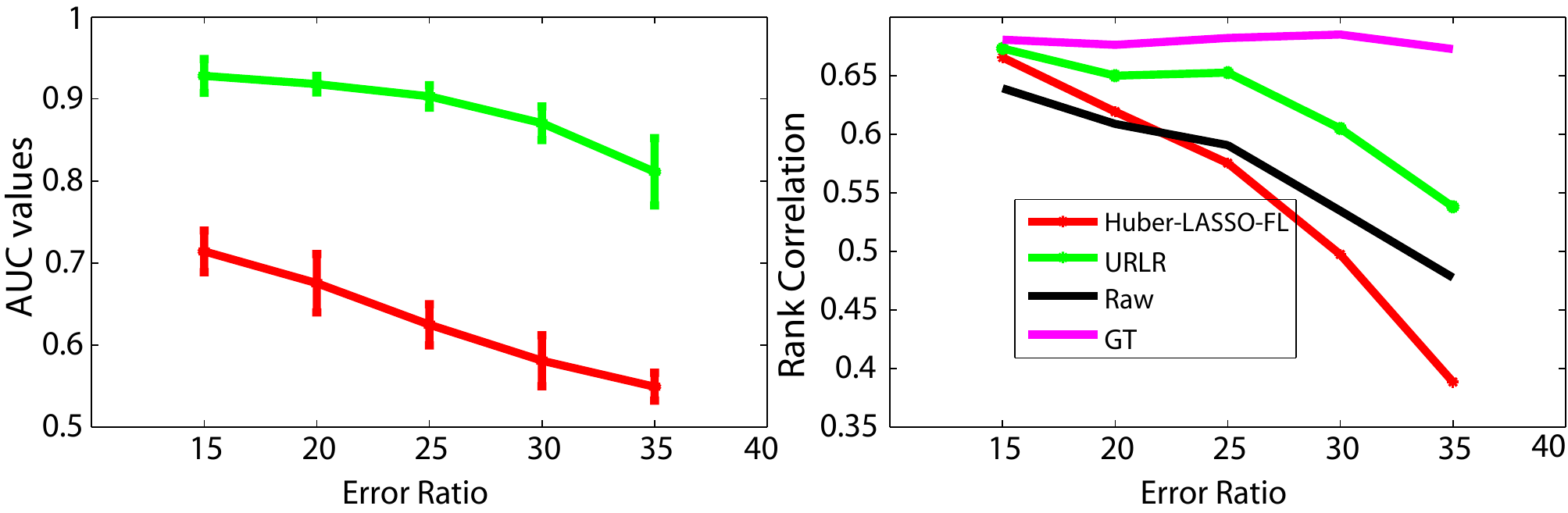}   
\protect\protect\caption{\label{fig:Error-Ratio.}Effect of error ratio. Left: outlier detection performance measured by area under ROC curve (AUC). Right: rank prediction performance measured by rank correlation.}
\end{figure}

\begin{figure}
\begin{centering}
\includegraphics[scale=0.45]{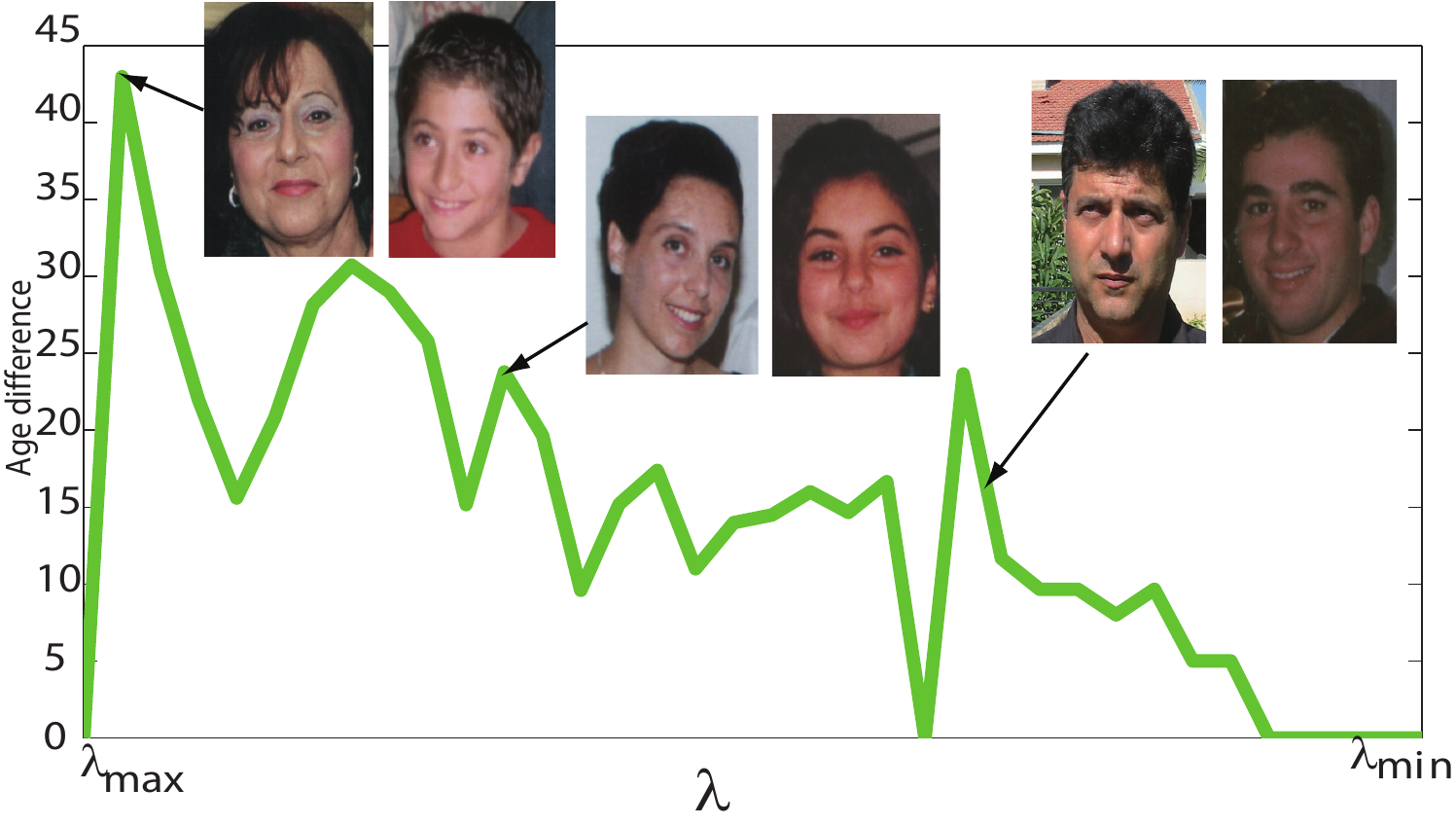}  
\par\end{centering}

\vspace{-0.3cm}
 \protect\protect\caption{\label{fig:regularizaitonpatheexample}Relationship between the pruning
order and actual age difference for URLR.}
\end{figure}

\noindent \emph{(3) Effects of error ratio.} We used the Unint.+Int.~error
model to vary the amount of random comparisons and simulate different
amounts of errors in $10$ sampled graphs from $300$ training images
and $2000$ unique sampled pairs from the training images. The pruning rate was fixed at $25\%$.
Figure \ref{fig:Error-Ratio.} shows that \emph{URLR} remains effective
even when the true error ratio reaches as high as $35\%$. This demonstrates that
although a sparse outlier model is assumed, our model can deal with
non-sparse outliers. It also shows that URLR consistently outperforms the alternative models especially when the error/outlier ratio is high. 

\noindent \textbf{What are pruned and in what order?} The effectiveness of the employed regularisation
path method for outlier detection can be examined as $\lambda$ decreases to produce a ranked list
for all pairwise comparisons according to the outlier probability. Figure
\ref{fig:regularizaitonpatheexample} shows the relationship between
the pruning order (i.e.~which pair is pruned first) and ground truth
age difference and illustrated by examples. It can be seen that overall
outliers with larger age difference tend to be pruned first. This
means that even with a conservative pruning rate, obvious outliers (potentially
causing more performance degradation in learning) can be reliably
pruned by our model.

\vspace{-0.3cm}
\section{Conclusions and Future Work}

We have proposed a novel unified robust learning to rank (URLR) framework
for predicting subjective visual properties from images and videos. The key advantage
of our method over the existing majority voting based approaches is
that we can detect outliers globally by minimising a global ranking
inconsistency cost. The joint outlier detection and feature based rank prediction
formulation also provides our model with an advantage over the conventional
robust ranking methods without features for outlier detection: it can be applied with a large number of candidates in comparison but a sparse sampling in crowdsourcing. 
The effectiveness of our model in comparison with state-of-the-art
alternatives has been validated on the tasks of image and video interestingness
prediction and predicting relative attributes for visual recognition. Its effectiveness for outlier detection has also been evaluated in depth in the human age estimation experiments. 

\red{By definition subjective visual properties (SVPs) are person-dependent. When our model is learned using  pairwise labels collected from many people, we are essentially learning consensus -- given a new data point the model aims to predict its SVP value  that can be agreed upon by most people. However, the predicted consensual SVP value  could be meaningless for a specific person when his/her taste/understanding of the SVP is completely different to that of most others. How to learn a person-specific SVP prediction model is thus part of the on-going work.} 
\red{Note that our model is only one of the possible solutions  to inferring global ranking from pairwise comparisons.  Other models exist. In particular, one  widely studied alternative  is  the  (Bradley-Terry-Luce (BTL) model~\cite{Hunter04mmalgorithms,AzariSoufiani_nips13_b,bradleyTerry}), which aggregates the ranking scores of pairwise comparisons to infer a global ranking by maximum likelihood estimation.  The BTL model is introduced to describe the probabilities of the possible outcomes when individuals are judged against one another in pairs \cite{Hunter04mmalgorithms}. It is primarily designed to incorporate contextual information in the global ranking model. We found that directly applying the BTL model to our SVP prediction task leads to much inferior performance because it does not explicitly detect and remove outliers. However, it is possible to integrate it into our framework to make it more robust against outliers and sparse labels whilst preserving its ability to take advantage of contextual information. }
Other new directions include extending the presented work to other applications where noisy pairwise labels exist, both in vision such as image denoising \cite{angularembedding}, iterative search and active learning of visual categories \cite{BiswasCVPR13}, and in other fields such as statistics and economics \cite{Hodgerank}.

\section*{Acknowledgements}
 This research of Jiechao Xiong was supported in part by National Natural Science Foundation of China: 61402019, and China Postdoctoral Science Foundation: 2014M550015.  The research of Yuan Yao was supported in part by National Basic Research Program of China under grant 2012CB825501 and 2015CB856000, as well as NSFC grant 61071157 and 11421110001. The research of Yanwei Fu and Tao Xiang was in part supported by a joint NSFC-Royal Society grant 1130360, IE110976 with Yuan Yao. 
 Yuan~Yao and Tao~Xiang are the corresponding authours.

\vspace{-0.3cm}
{\small{}{} \bibliographystyle{ieeetr}
\bibliography{ref-phd1}
 }

\vspace{-0.6cm} 
\begin{IEEEbiography}[{\includegraphics[width=1in,height=1.25in,clip,keepaspectratio]{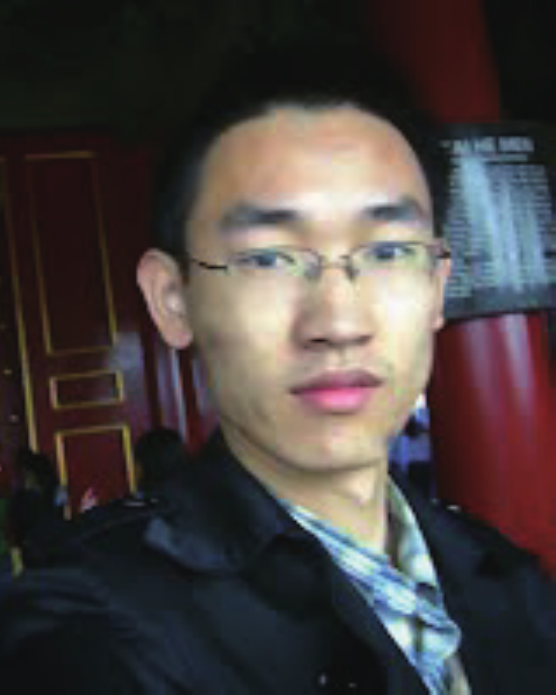}}]{Yanwei Fu} received the PhD degree from Queen Mary University of London in 2014, and the MEng degree in the Department of Computer Science \& Technology at Nanjing University in 2011, China. His research interest is large-scale image and video understanding, graph-based machine learning algorithms, robust ranking and robust learning to rank.
\end{IEEEbiography}
\vspace{-1cm}
\begin{IEEEbiography}
[{\includegraphics[width=1in,height=1.25in,clip,keepaspectratio]{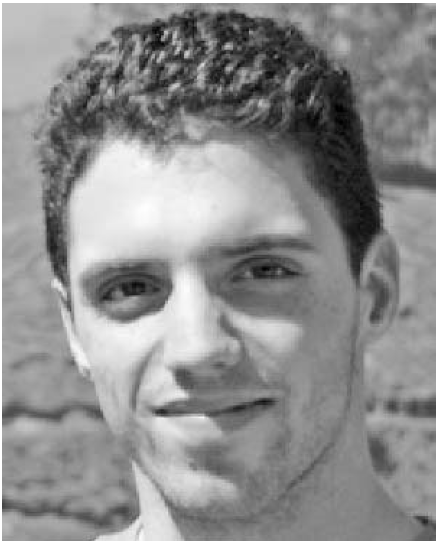}}]{Timothy M. Hospedales} received  the  PhD degree in neuroinformatics from the University of Edinburgh in 2008. He is currently a lecturer (assistant professor) of computer science at Queen Mary University of London. His research interests include probabilistic modelling and machine learning applied variously to problems in computer vision, data mining, interactive learning, and neuroscience. He has published more than 30  papers  in  major  international journals and conferences. He is a member of the IEEE.
\end{IEEEbiography}

\vspace{-1cm} 
\begin{IEEEbiography}[{\includegraphics[width=1in,height=1.25in,clip,keepaspectratio]{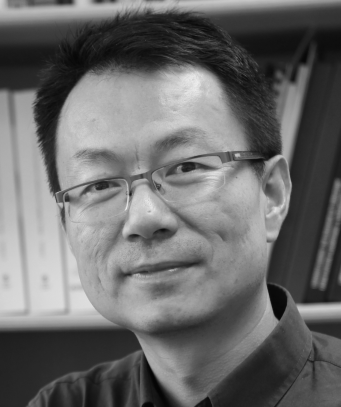}}]{Tao Xiang} received the PhD degree in electrical and computer engineering from the National University of Singapore in 2002. He is currently a reader (associate professor) in the School of Electronic Engineering and Computer Science, Queen Mary University of London. His research interests include computer vision and  machine learning. He has published over 100 papers in international journals and conferences and co-authored a book, Visual Analysis of Behaviour: From Pixels to Semantics.
\end{IEEEbiography}

\vspace{-1cm}
\begin{IEEEbiography}
[{\includegraphics[width=1in,height=1.25in,clip,keepaspectratio]{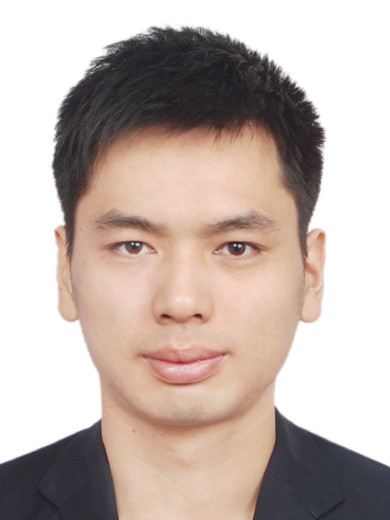}}]{Jiechao Xiong} is currently pursuing the Ph.D. degree in statistics in BICMR $\&$ School of Mathematical Science, Peking University, Beijing, China. His research interests include statistical learning, data science and topological and geometric methods for high-dimension data analysis.
\end{IEEEbiography}

\vspace{-1cm} 
\begin{IEEEbiography}[{\includegraphics[width=1in,height=1.25in,clip,keepaspectratio]{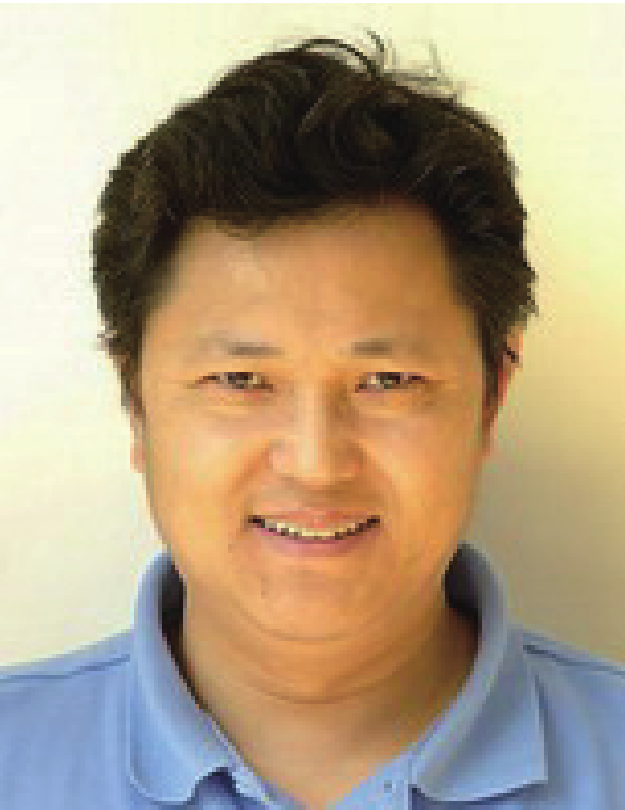}}]{Shaogang Gong}  is Professor of Visual Computation at Queen Mary University of London, a Fellow of the Institution of Electrical Engineers and a Fellow of the British Computer Society. He received his D.Phil in computer vision from Keble College, Oxford University in 1989. His research interests include computer vision, machine learning and video analysis.
\end{IEEEbiography}

\vspace{-1cm} 
\begin{IEEEbiography}[{\includegraphics[width=1in,height=1.25in,clip,keepaspectratio]{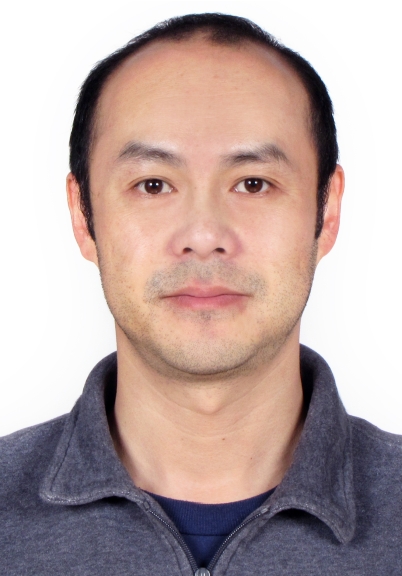}}]{Yizhou Wang}  is a Professor in the Department of Computer Science at Peking University, Beijing, China. He is a vice director of Institute of Digital Media at Peking University, and the director of New Media Lab of National Engineering Lab of Video Technology. He received his Ph.D. in Computer Science from University of California at Los Angeles (UCLA) in 2005.  Dr. Wang's research interests include computational vision, statistical modeling and learning, pattern analysis, and digital visual arts.

\end{IEEEbiography}

\vspace{-1cm} 
\begin{IEEEbiography}[{\includegraphics[width=1in,height=1.25in,clip,keepaspectratio]{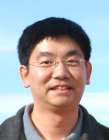}}]{Yuan Yao} received his Ph.D. in mathematics from the University of California, Berkeley, in 2006. Since then he has been with Stanford University and in 2009, he joined the School of Mathematical Sciences, Peking University, Beijing, China, as a professor of statistics. His current research interests include topological and geometric methods for high dimensional data analysis and statistical machine learning, with applications in computational biology, computer vision, and information retrieval.
\end{IEEEbiography}

\section*{Supplementary Material}

\noindent Thanks for the excellent questions from the anonymous reviewers
of our TPAMI submission. In answering their questions, we found some
details and insights of our framework which have been overlooked before.
Due to the page limits of our journal version, we use this document
to further explain the details and insights and help our readers better
understand our work.
\begin{enumerate}
\item Further, the proposed approach doesn't seem to truly get to the bottom
of why subjective properties are tricky namely that two people might
actually have a different understanding of the property. While the
authors do refer to such possible disagreements in the introduction,
the proposed method doesn't seem to consider this possibility. In
other words, how does it make sense to consider a single global order
when such an order might be unattainable since person A's \textquotedbl{}interestingness\textquotedbl{}
will differ from person B's?

\textit{\textcolor{blue}{This is a very good question. Indeed, since
the properties are subjective, they are by definition person-dependent.
However, in most applications when we learn a SVP prediction model
using pairwise labels collected from many different annotators, we
are modeling consensus. In other words, the model essentially aggregates
the understandings of different people regarding a certain SVP so
that the predicted SVP for an unseen data point can be agreed upon
by most people. For example, in the case of video interestingness,
YouTube may want to predict the interestingness of a newly uploaded
video so as to decide whether or not to promote it. Such a prediction
obviously needs to be based on consensus from the majority of the
YouTube viewers regarding what defines interestingness. However, collecting
consensus can be expensive; the proposed model in this paper thus
aims to infer the consensus from as few labels as possible. }}

\textit{\textcolor{blue}{It is also true that for a specific person,
he/she would prefer a SVP prediction model that is tailor-made for
his/her own understanding of the SVP, i.e. a person-specific prediction
model. Such a model needs to be learned using his/her pairwise labels
only. For example, YouTube could recommend different videos for different
registered users when they log in, if they provide some pairwise video
interestingness labels for learning such a model (at present, this
is done based on some simple rules from the viewing history of the
user). This also has its own problem - it is much harder to collect
enough labels from a single person only to learn the prediction model.
There are solutions, e.g. categorising the users into different groups
so that the labels from people of the same group can be shared. However
this is beyond the scope of this paper and is being considered as
part of ongoing work. }}

\textit{\textcolor{blue}{We have provide a discussion on this problem
in Section 5 in the revised manuscript (Page 14). }}

\item It feels a little bit unsatisfying that the method requires we pick
a fixed ratio of outliers. This would be more ok if the ratio can
be automatically computed from the data somehow.

\textit{\textcolor{blue}{Indeed, the pruning rate is a free parameter
of the proposed model (in fact, the only free parameter) that has
to set manually. As discussed in the beginning of Section 3.3, most
existing outlier detection algorithms have a similar free parameter
determining how aggressive the algorithm needs to be for pruning outliers.
Automated model selection critieria such as BIC and AIC could be considered.
However, as pointed out by {[}49{]}, they are often unstable for the
outlier detection problem with pairwise labels. We have carried out
experiments to show that when BIC or AIC is employed, the selected
model failed to detect meaningful outliers. Since a related comment
is given by Reviewer 3, please refer to the Response Point 2 to Reviewer
3 for detailed experiment results and analysis on the alternative
outlier detection methods including BIC. It is also worth pointing
out that our results on the effect of the pruning rate show that the
proposed model remains effective given a wide range of pruning rate
values (see Fig. 3, 5, 9 and 10). }}

\textit{\textcolor{blue}{We have now added a footnote in Section 3.3
to discuss why an automated model selection criterion such as BIC
is not adopted. }}

\item \textcolor{black}{I think cases of Raw performing similarly or better
than MajVot1/ 2 should be explained in a little more detail, i.e.
an intuition for such outcomes should be given.}

\textit{\textcolor{blue}{Thanks for the suggestion. Indeed, our results
on both image and video interestingness experiments show that Raw
performs similarly to majority voting. There is an intuitive explanation
for that. When a pair of data points A and B receive multiple votes/labels
of different pairwise orders/ranks, these multiple labels are converted
into a single label corresponding to the order that receives the most
votes. Since only one of the two orders is correct (either A>B or
B>A), there are two possibilities: the majority voted label is correct,
or incorrect, i.e. an outlier. In comparison, using Raw, all votes
count, so the outlying votes would certainly having a negative effect
on the learned prediction model, so would the correct votes/labels.
Now let us consider which method is better. The answer is it depends
on the outlier/error ratio of the labels. If the ratio is very low,
majority voting will get rid of almost all the outlying votes; MajVot
would thus be advantageous over Raw which can still feel the negative
effects of the outliers. However, when the ratio gets bigger, it becomes
possible that the outlying label becomes the winning vote. For example,
if A>B is correct, and received 2 votes and A<B is incorrect and received
3 outlying votes. Using Raw, those 2 correct votes still contribute
positively to the model, whilst using MajVot, their contribution disappears
and the negative impact of the outlying votes is amplified. Therefore,
one expects that when MajVot makes more and more mistakes, its performance
will get closer to that of Raw, until it reaches a tipping point where
Raw starts to get ahead. }}

\textit{\textcolor{blue}{We have added a brief discussion on this
in Section 4.1. on Page 9. }}

\item In Figure 3, why does the Kendall tau distance start to increase as
the pruning rate increases, after 55\% for URLR?

\textit{\textcolor{blue}{Higher Kendall tau distance means worse prediction.
Figure 3 (right) thus shows that our URLR's performance is improved
when more and more outliers are pruned in the beginning; then after
more than 55\% of pairs are pruned, its performance starts to decrease.
This result is expected: at low pruning rates, most of the pruned
pairs are outliers; the model therefore benefits. Since the percentage
of outliers would almost certainly be lower than 50\%, when the pruning
rate reaches 55\%, most of the outliers have been removed, and the
algorithm start to remove the correctly labelled pairs. With less
and less correct labels available to learn the model, the performance
naturally would decrease - when pruning rate gets close to 100\%,
it would not be possible to learn a meaningful model; the Kendall
tau distance would thus shoot up. }}

\textit{\textcolor{blue}{We have now added a sentence on Page 10 to
give an explanation to this phenomenon. }}

\item \textcolor{black}{Page 2 line 44 \textquotedblleft For example, Figure
1 \dots{} \textquotedblright{} the authors try to argue that examples
shown in Figure 1 are outliers. I don\textquoteright t quite agree.
Authors are trying to study subjective attributes. These are good
examples of subjective versus objective attributes. This doesn't seem
to be about outliers vs. not. In fact, one source of outliers other
than malicious workers is global (in)consistency, which is not mentioned
here. The authors could draw from the concrete example of Figure 2.}

\textit{\textcolor{blue}{This is a very good point. It is certainly
worthwhile to clarify the definition of outlier in the context of
subjective visual property (SVP). In particular, since by definition
a SVP is subjective, defining outlier, even making the attempt to
predict SVP is self-contradictory - one man's meat is another man's
poison. However, there is certainly a need for learning a SVP prediction
model, hence this paper. This is because when we learn the model from
labels collected from many people, we essentially aim to learn the
consensus, i.e. what most people would agree on (please see our Response
Points 1 for more discussion on this). Therefore, Figure 1(a) can
still be used to illustrate this outlier issue in SVP annotation,
that is, you may have most of the annotators growing up watching Sesame
Street thus consciously or subconsciously consider the Cookie Monster
to be more interesting than the Monkey King; their pairwise labels/votes
thus represent the consensus. In contrast, one annotator who is familiar
with the stories in Journey to the West may choose the opposite; his/her
label is thus an outlier under the consensus. We have reworded the
relevant text on Page 2 to avoid confusion.}}

\item A baseline to compare to might be to feed all individual constraints
(without majority vote) to a rankSVM. SVMs already allow for some
slack. So I would be curious to know if that takes care of some of
the outliers already.

\textit{\textcolor{blue}{Thanks. In fact, we do have one set of results
on this. Specifically, in Sec 4.2 ``Video interestingness prediction'',
as explained under ``Experimental settings'', we employed rankSVM
model to replace Eq (9). Therefore, the model denoted as 'Raw' in
this experiment is exactly the suggested baseline of feeding all constraints
to a rankSVM. As shown in Fig. 5(a), the model is at par with Maj-Vot-1
but worse then the two global outlier detection methods Huber-LASSO-FL
and our URLR. This result suggests that rankSVM does have some ability
to cope with outliers. However, we are not sure this is due to the
slack variables of rankSVM. This is because the slack variables are
introduced to account for data noise \cite{ranking_books} which is
different from the outliers in the pairwise data. }}

\item For the scene and pubfig image dataset, the relative attribute prediction
performance can only be evaluated indirectly by image classification
accuracy with the predicted relative attributes as image representation.\textquotedbl{}
> Why is that? Can't you compute attribute prediction performance
on a held out set of annotated pairs? Or is the concern that since
the pairs may be noisily annotated, one can not think of them as GT?
But is that not an issue with interestingness then? Please clarify
in rebuttal.

\textit{\textcolor{blue}{Thanks for this question. We stated in footnote
9 that ``Collecting ground truth for subjective visual properties
is always problematic. Recent statistical theories {[}61{]}, {[}19{]}
suggest that the dense human annotations can give a reasonable approximation
of ground truth for pairwise ranking. This is how the ground truth
pairwise rankings provided in {[}4{]} and {[}5{]} were collected.''
So for image and video interestingness as well as the age dataset,
(dense) enough pairwise comparisons are available to give a reasonable
approximation of the groundtruth. However, this is not the case for
scene and pubfig image dataset: the collected pairs are much more
sparse and cannot be used as an approximation to the groundtruth.
In short, it is because they are too sparse rather than too noisy.
}}\\
\textit{\textcolor{blue}{In contrast, the indirect evaluation metric
of downstream classification accuracy has clear unambiguous groundtruth,
and directly depends on relative attribute prediction accuracy. So
this evaluation is preferred.}}

\item Related Work: The Bradley-Terry-Luce (BTL) model is the standard model
for computing a global ranking from pairwise labels. It should be
mentioned in the related work. See {[}52{]} or Hunter, D. R. (2004).
MM algorithms for generalized BradleyTerry models. Annals of Statistics.
Experiments: I would expect additional comparisons to state-of-the-art
(BTL or SVM-rank aggregation {[}52{]}). In particular the Bradley-Terry-Luce
(BTL) model is extremely widely used and more robust to noise than
LASSO based approaches {[}52{]}. E.g. \textquotedbl{}Generalized Method-of-Moments
for Rank Aggregation\textquotedbl{} or \textquotedbl{}Efficient Bayesian
Inference for Generalized Bradley-Terry Models\textquotedbl{} provide
code for inference in BTL models. Such a method leads to a global
ranking, which could be used to train an SVM. Alternatively, it can
be used to find pairwise rankings that disagree with the obtained
global ranking. These could be removed as outliers and a rank-SVM
trained from the remaining pairwise labels. Such an experiment should
be included as an additional state-of-the-art comparison in the updated
version of the manuscript.

\textit{\textcolor{blue}{Thanks for the suggestion. Indeed, the Bradley-Terry-Luce
(BTL) model is a very relevant global ranking model. We have now studied
it carefully and made connections to the proposal URLR model. We also
carried out new experiments to evaluate the BLT model for our Subjective
Visual Property (SVP) prediction task. }}

\textit{\textcolor{blue}{More specifically, the BTL model is a probabilistic
model that aggregates the ranking scores of pairwise comparisons to
infer a global ranking by maximum likelihood estimation. It is closely
related to the proposed global ranking model; yet it also has some
vital differences. Let's first look at the connection. The main pairwise
ranking model of Huber-LASSO used in this paper is a linear model
(see Eq (10) and Eq (12)), which is }}

\textit{\textcolor{blue}{
\begin{equation}
y_{ij}=\theta_{i}-\theta_{j}+\gamma_{ij}+\varepsilon_{ij}\label{eq:ranking_model}
\end{equation}
}}

\textit{\textcolor{blue}{In statistics and psychology \cite{Hodgerank,bredleyTerryJMLR2006,lars,rank_analysis_bradleyTerry},
such a linear model can be extended to a family of generalised linear
models when only binary comparisons are available for each pair $(i,j)$,
i.e. either $i$ is preferred to $j$ or vice versa. In these generalised
linear models, one assumes that the probability of pairwise preference
is fully determined by a linear ranking/rating function in the following,}}

\textit{\textcolor{blue}{
\[
\pi_{ij}=\mathrm{Prob}\left\{ i\: is\, preferred\, over\, j\right\} =\Phi\left(\theta_{i}-\theta_{j}\right)
\]
where $\Phi:\mathbb{R}\rightarrow\left[0,1\right]$ can be chosen
as any symmetric cumulated distributed function. }}

\textit{\textcolor{blue}{Different choices of $\Phi$ lead to different
generalised linear models. In particular, two choices are worth mentioning
here:}}
\begin{itemize}
\item \textit{\textcolor{blue}{Uniform model, 
\begin{equation}
y_{ij}=2\pi_{ij}-1\label{eq:linear}
\end{equation}
This model is equivalent to use $y_{ij}=1$ if $i$ is preferred to
$j$ and $y_{ij}=-1$ otherwise in linear model. This model is used
in this work to derive our URLR model. }}
\item \textit{\textcolor{blue}{Bradley-Terry-Luce (BTL) model, 
\begin{equation}
y_{ij}=\mathrm{log}\frac{\pi_{ij}}{1-\pi_{ij}}\label{eq:BTL}
\end{equation}
}}
\end{itemize}

\textit{\textcolor{blue}{So by now, it is clear that both our URLR
and BTL generalise the linear model in Huber-LASSO. They differ in
the choice of the symmetric cumulated distributed function $\Phi$.}}

\textit{\textcolor{blue}{Although both of them are generalised from
the same linear model, they are developed for very different purposes.
The BTL model is introduced to describe the probabilities of the possible
outcomes when individuals are judged against one another in pairs
\cite{Hunter04mmalgorithms,rank_analysis_bradleyTerry}. It is primarily
designed to incorporate contextual information in the global ranking
model. For instance, in sports applications, it can be used to account
for the home-field advantage and ties situations \cite{bredleyTerryJMLR2006,bradleyTerry}.
In contrast, our framework tries to detected outliers in the pairwise
comparisons and cope with the sparse labels. Consequently, from Eq
(1) onwards, we introduce the outlier variable to model the outiers
explicitly and introduce low-level feature variable to enhance our
model's ability to detect outliers given sparse labels. None of these
is in the BLT model, which means that it may not be suitable given
sparse pairwise comparisons with outliers. }}

\textit{\textcolor{blue}{To verify this, we took the suggestion by
Reviewer 3 and employed the matlab codes from the website of \cite{bradleyTerry}
\textquotedbl{}Efficient Bayesian Inference for Generalized Bradley­Terry
Models\textquotedbl{} to carry out experiments. The results on image
interestingness prediction are compared in Fig \ref{fig:Comparing-BTL-model}.
It shows that the performance of BTL is much worse than the other
alternatives. Similar results were obtained on video interestingness
prediction and age estimation. }}

\begin{figure}
\begin{centering}
\includegraphics[scale=0.5]{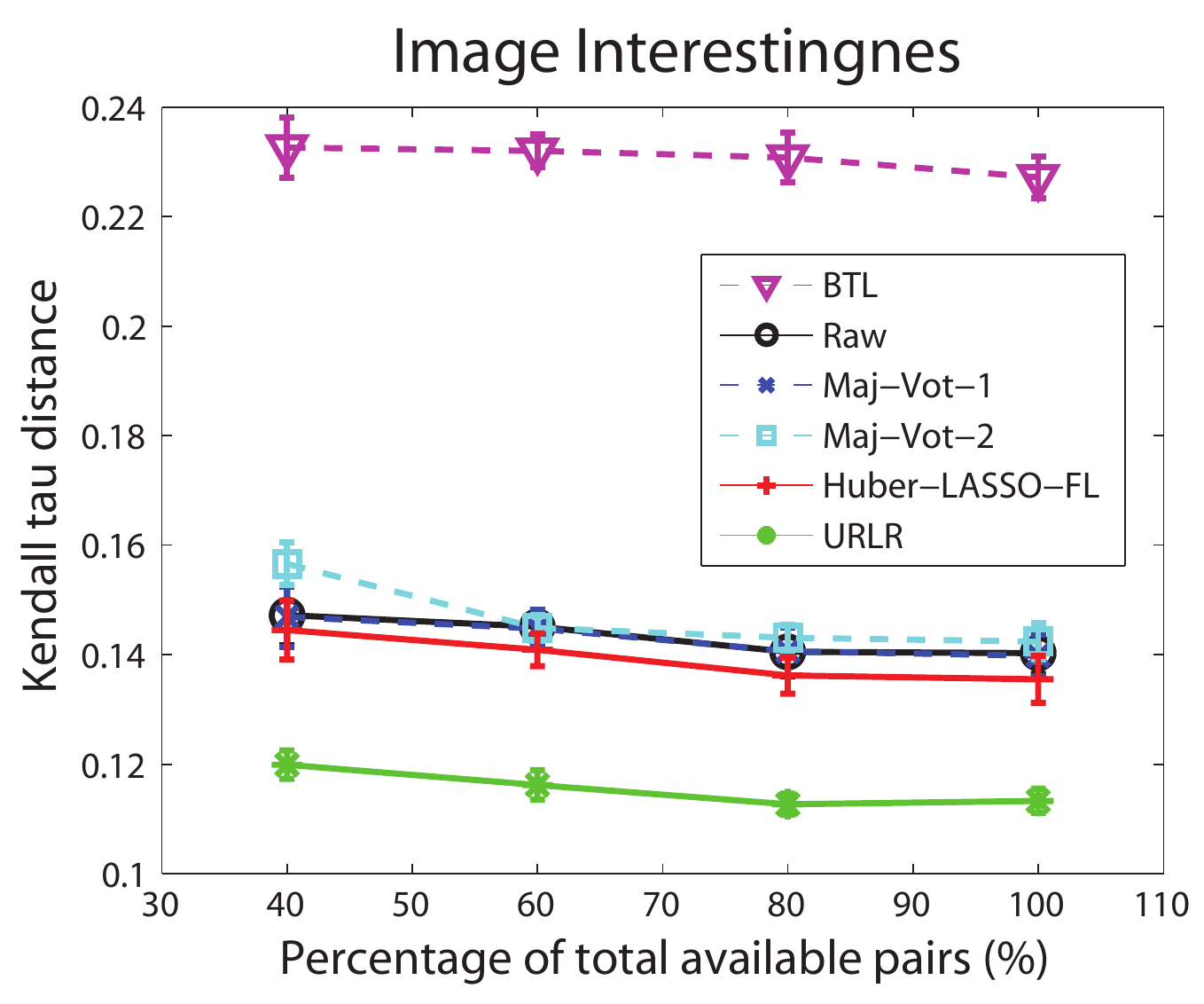}
\par\end{centering}

\protect\caption{\label{fig:Comparing-BTL-model}Comparing the BTL model with our model
on image interestingness prediction}
\end{figure}

\textit{\textcolor{blue}{As explained above, it is actually not fair
to compare the BTL model to the other models because BTL was not designed
for outlier detection and could not cope with the amount of outliers
and the level of spareness in our SVP data. We therefore decide not
to include the new results in the revised manuscript. However, from
our analysis above, it is also clear that we could use the BTL model
(Eq (\ref{eq:BTL})) to generalise the linear model in place of the
uniform model, and use it in our outlier detection framework. In this
way, we can have the better of both worlds: the ability of BTL to
incorporate contextual information such as the home-filed advantage
in sports can also be taken advantage of in our framework whilst preserving
our model's strength on robustness against outliers and sparse labels.
However, this is probably beyond the scope of this paper and is better
left to the future work. In the revised manuscript, we have now added
the following paragraph in Section 5, where we discuss that BTL is
an alternative model that can be integrated into our framework as
part of the future work.}}
\begin{quote}
\textit{\textcolor{blue}{``Note that our model is only one of the
possible solutions to inferring global ranking from pairwise comparisons.
In particular, one widely studied alternative  is the (Bradley-Terry-Luce
(BTL) model {[}61,62,63{]}, which aggregates the ranking scores of
pairwise comparisons to infer a global ranking by maximum likelihood
estimation. The BTL model is introduced to describe the probabilities
of the possible outcomes when individuals are judged against one another
in pairs {[}61{]}. It is primarily designed to incorporate contextual
information in the global ranking model. We found that directly applying
the BTL model to our SVP prediction task leads to much inferior performance
because it does not explicitly detect and remove outliers. However,
it is possible to integrate it into our framework to make it more
robust against outliers and sparse labels whilst preserving its ability
to take advantage of contextual information.''}}
\end{quote}
\item 3.3 Regularization path. On the one hand the authors say that \textquotedbl{}Setting
a constant \textgreek{l} value independent of dataset is far from
optimal because the ratio of outliers may vary for different crowdsourced
datasets\textquotedbl{}, but using the regularization path this is
exactly what is done in the end. It is true that the experiments show
that the proposed method is fairly robust w.r.t. the outlier ratio.
Nonetheless, I would like to see an experiment using a (modified)
BIC for selecting the outlier ratio. This would be a valuable extension
over the ECCV work.

\textit{\textcolor{blue}{Thanks. As discussed in the beginning of
Section 3.3, most existing outlier detection algorithms have a similar
free parameter as \textgreek{l} to determine how aggressive the algorithm
needs to be for pruning outliers. Automated model selection critieria
such as BIC and AIC could be considered. However, as pointed out by
{[}49{]}, they are often unstable for the outlier detection problem
with pairwise labels. }}

\textit{\textcolor{blue}{We have evaluated alternative methods including
the modified BIC and AIC for image and video interestingness prediction.
The results suggest those automated models such as AIC and BIC failed
to identify any outliers - they prefer the model that include all
input pairwise comparisons. To find out why it is the case, we carried
out a controlled experiment using synthetic data to investigate how
different factors affect the performance of different methods for
determining the outlier ratio. Specifically, we compare , BIC and
with our Regularization Path model.}}

\textbf{\textit{\textcolor{blue}{Experiments design.}}}\textit{\textcolor{blue}{{}
we use a complete graph $G$ with $30$ nodes. Our framework is simplified
into the following ranking model,}}

\textit{\textcolor{blue}{
\[
Y_{ij}=\theta_{i}-\theta_{j}+\gamma_{ij}+\varepsilon_{ij}
\]
}}

\textit{\textcolor{blue}{Let $\theta\sim U\left(-1,1\right)$, $\varepsilon_{ij}\sim\mathcal{N}(0,\sigma^{2})$
and $\gamma_{ij}=\pm L$. We simulate the outlier pairs by randomly
sampling, that is, each pair's true ranking is reversed (i.e. becoming
an outlier/error) with a probability $p$ which will determine the
outlier ratio. The magnitude of outliers in relation to that of the
noise is another factor which could potentially affect the performance
of different methods on outlier detection. So we define the outlier-noise-ratio
$ONR:=L/\sigma$, where $\sigma=0.1$ in our experiment and $L$ is
varied in our experiment to give different ONR values. }}

\textbf{\textit{\textcolor{blue}{Evaluation protocols}}}\textit{\textcolor{blue}{{}
}}\textbf{\textit{\textcolor{blue}{and results}}}\textit{\textcolor{blue}{.
We first compare three methods that require the manual setting of
a free parameter corresponding to the outlier ratio. These include
our formulation (Eq (8)) with Regularization Path (i.e. the proposed
model), IPOD hard-threshold \cite{penalizedlasso}}}%
\footnote{\textit{\textcolor{blue}{Strictly speaking, IPOD hard-threshold is
not a Lasso solver, since it replaced the soft-thresholding with hard-thresholding.
However, for comparison convenience, we still compare it with our
RP.}}%
}\textit{\textcolor{blue}{{} with Regularization Path, and our formulation
with orthogonal matching pursuit \cite{OrthogonalMatchingPursuit}.
Using our model with Regularization Path, $\lambda$ is decreased
from $\infty$ to 0 and the graph edges are order according to how
likely it corresponds to an outlier. The top $p\%$ edge set $\Lambda_{p}$
are detected as outliers. By varying $p$, ROC (receiver-operating-characteristic)
curve can be plotted and AUC (area under the curve) is computed. Similarly,
IPOD hard-threshold can also be solved using the same Regularization
Path strategy. And orthogonal matching pursuit can be used to solve
our formulation for outlier detection in place of Regularization Path.
As shown in Figure \ref{fig:Comparing-methods-of-RP}, the results
of our formulation with Regularization Path are consistently better
than those of IPOD hard-threshold + Regularization Path and our formulation
+ orthogonal matching pursuit. Specifically, it shows that (1) when
there are small portions of outliers, all the methods can reliably
prune most of outliers; (2) in all experiments, IPOD-hard threshold
and orthogonal matching pursuit have similar performance, whilst our
formulation + Regularization Path is consistently better than the
other alternatives, especially when there are large portions of outliers
(high values of $p$); (3) the higher the ONR, the better performance
of outlier detection for all three methods. }}

\textcolor{blue}{}
\begin{figure*}[t]
\centering{}\textcolor{blue}{\includegraphics[scale=0.45]{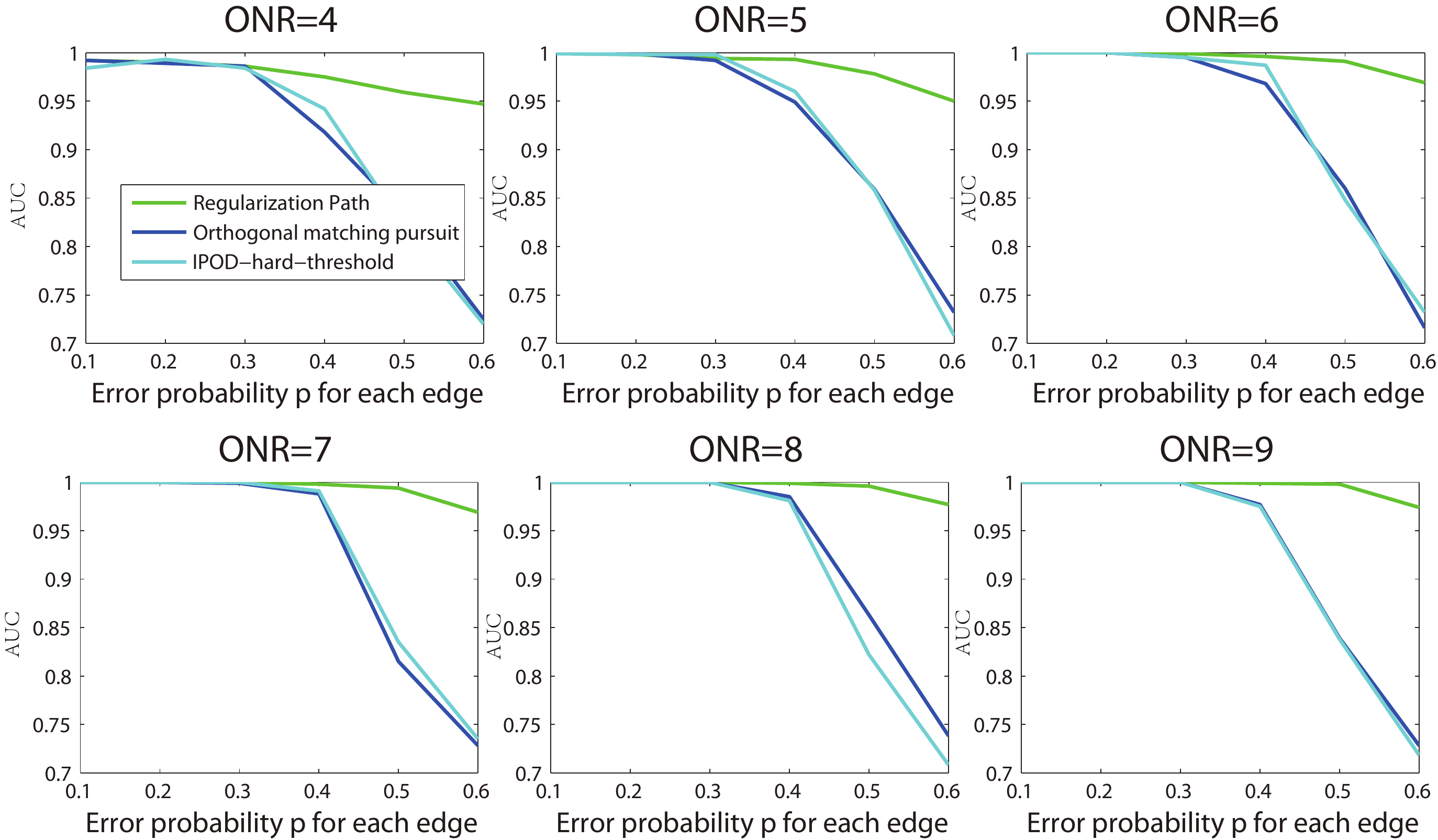}\protect\caption{\label{fig:Comparing-methods-of-RP}Effects of outlier/error probability
(p) and outlier-noise ratio (ONR) on our formulation + Regulaization
Parth (denoted as Regulaization Parth), IPOD-hard threshold + Regulaization
Parth and our formulation + Orthogonal matching pursuit.}
}
\end{figure*}

\textcolor{blue}{{} }
\begin{table*}[tbh]
\begin{centering}
\textcolor{blue}{\tiny{}}%
\begin{tabular}{|c|c|c|c|c|c|c|}
\hline 
 & \textcolor{blue}{\tiny{}ONR=4} & \textcolor{blue}{\tiny{}ONR=5} & \textcolor{blue}{\tiny{}ONR=6} & \textcolor{blue}{\tiny{}ONR=7} & \textcolor{blue}{\tiny{}ONR=8} & \textcolor{blue}{\tiny{}ONR=9}\tabularnewline
\hline 
\hline 
\textcolor{blue}{\tiny{}p=0.1} & \textcolor{blue}{\tiny{}0.002/0} & \textcolor{blue}{\tiny{}0.494/0.012} & \textcolor{blue}{\tiny{}1/0.003} & \textcolor{blue}{\tiny{}1/0.026} & \textcolor{blue}{\tiny{}1/0.025} & \textcolor{blue}{\tiny{}1/0.031}\tabularnewline
\hline 
\textcolor{blue}{\tiny{}p=0.2} & \textcolor{blue}{\tiny{}0/0} & \textcolor{blue}{\tiny{}0/0} & \textcolor{blue}{\tiny{}0.3/0.016} & \textcolor{blue}{\tiny{}0.9/0.05} & \textcolor{blue}{\tiny{}1/0.064} & \textcolor{blue}{\tiny{}1/0.037}\tabularnewline
\hline 
\textcolor{blue}{\tiny{}p=0.3} & \textcolor{blue}{\tiny{}0/0} & \textcolor{blue}{\tiny{}0/0} & \textcolor{blue}{\tiny{}0/0} & \textcolor{blue}{\tiny{}0/0} & \textcolor{blue}{\tiny{}0/0} & \textcolor{blue}{\tiny{}0.5/0.06}\tabularnewline
\hline 
\textcolor{blue}{\tiny{}p=0.4} & \textcolor{blue}{\tiny{}0/0} & \textcolor{blue}{\tiny{}0/0} & \textcolor{blue}{\tiny{}0/0} & \textcolor{blue}{\tiny{}0/0} & \textcolor{blue}{\tiny{}0/0} & \textcolor{blue}{\tiny{}0/0}\tabularnewline
\hline 
\textcolor{blue}{\tiny{}p=0.5} & \textcolor{blue}{\tiny{}0/0} & \textcolor{blue}{\tiny{}0/0} & \textcolor{blue}{\tiny{}0/0} & \textcolor{blue}{\tiny{}0/0} & \textcolor{blue}{\tiny{}0/0} & \textcolor{blue}{\tiny{}0/0}\tabularnewline
\hline 
\textcolor{blue}{\tiny{}p=0.6} & \textcolor{blue}{\tiny{}0/0} & \textcolor{blue}{\tiny{}0/0} & \textcolor{blue}{\tiny{}0/0} & \textcolor{blue}{\tiny{}0/0} & \textcolor{blue}{\tiny{}0/0} & \textcolor{blue}{\tiny{}0/0}\tabularnewline
\hline 
\end{tabular}
\par\end{centering}{\tiny \par}

\textcolor{blue}{\protect\caption{\label{tab:TPR-and-FPR}The outlier detection results of our formulation
+ BIC. The results are presented as TPR/FPR. The error probability
and ONR are: $p\in\left[0.1,0.6\right]$ and $ONR\in\left[4,9\right]$
respectively.}
}
\end{table*}

\textcolor{blue}{}\textit{\textcolor{blue}{In contrast, BIC utilises
the relative quality and likelihood functions of statistical models
themselves to determine a fixed $\lambda$. Therefore, the true positive
rate (TPR) and false positive rate (FPR) for BIC are reported. The
results are listed in Table \ref{tab:TPR-and-FPR}. It shows that
when using our formulation with BIC, only when there are very small
portions of outliers and the outlier-noise-ratio is extremely high,
BIC can reliably prune most of outliers. Otherwise, it tends to consider
all pairs inliers. As mentioned above, using BIC in place of Regularization
Path also leads to no outliers being pruned in our SVP prediction
experiments. This thus suggests that the real outlier ratio (roughly
corresponds to $p$=0.2, see Response Point 10 to Reviewer 2) and/or
outlier-noise-ratio (ONR) are too high for BIC to work. }}

\textit{\textcolor{blue}{Due to the space constraint, we could not
include all these results and analysis in the revised manuscript.
On Page 6, we have now added a footnote (Footnote 3) to refer the
readers to find additional results and discussion on this outlier
ratio problem in the project webpage at http://www.eecs.qmul.ac.uk/\textasciitilde{}yf300/ranking/index.html. }}

\item Page 9, Col. 2, Line 52: The authors talk about global image features
(GIST), but Page 8, Line 45 indicates that the ground truth annotations
such as \textquotedblleft central object\textquotedblright , etc.
were used. Using the complete ground truth annotation seems to be
problematic, as it also contains an attribute \textquotedbl{}is interesting\textquotedbl{}
and others such as \textquotedbl{}is aesthetic\textquotedbl{} and
\textquotedbl{}is unusual\textquotedbl{}. When using this ground truth,
I believe such labels should be excluded and only content attributes
used. (such as: indooroutdoor, contains a person, etc.).

\textit{\textcolor{blue}{Thanks for the suggestion. We have updated
this experiment as suggested. Specifically, we first examined how
each of the 932 attribute features are correlated to the groundtruth
interestness value of each image. Figure \ref{fig:Kendall-tau-correlations}.
shows that (1) only small number of these attribute features have
strong correlation with the interestingness value. (2) the histogram
of kendall tau correlations}}%
\footnote{\textit{\textcolor{blue}{Note that here, we employ kendall tau correlation
rather than the Spearman correlation (Spearman correlation of ``is
interesting'' vs. groundtruth is 0.63 as reported in \cite{imginterestingnessICCV2013})
since Spearman correlation is much more sensitive to error and discrepancies
in data and Kendall tau correlation \cite{kendtall1990} generally
have better statistical properties.}}%
}\textit{\textcolor{blue}{{} of all features is roughly Gaussian as shown
in Fig. \ref{fig:Kendall-tau-correlations}(right).}}

\textit{\textcolor{blue}{}}
\begin{figure*}
\begin{centering}
\includegraphics[scale=0.4]{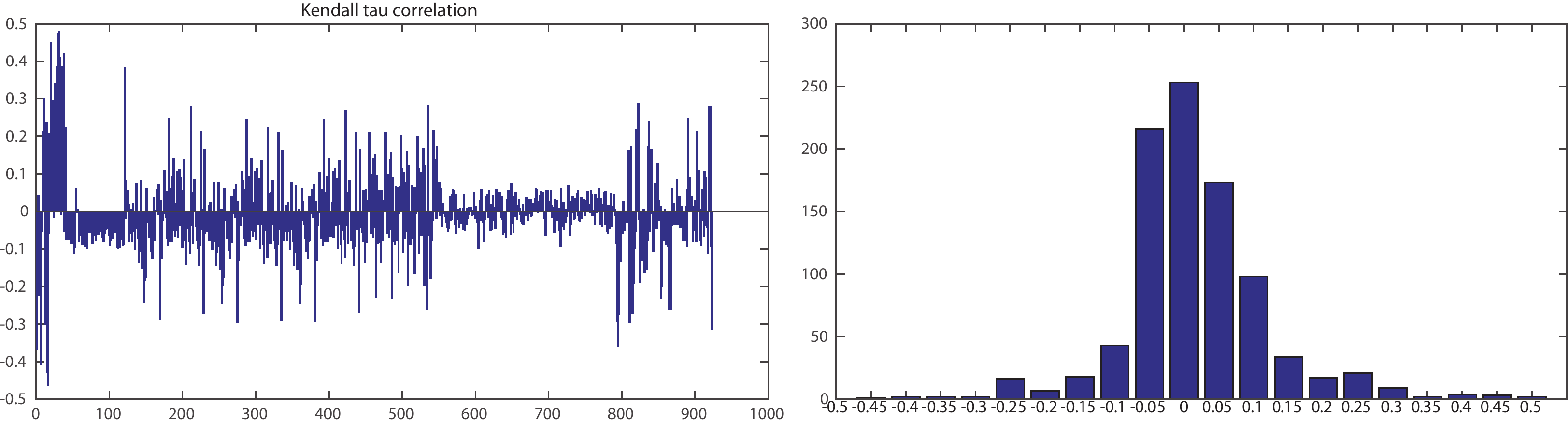}
\par\end{centering}

\textit{\textcolor{blue}{\protect\caption{\label{fig:Kendall-tau-correlations}Kendall tau correlations of each
feature dimension with the ground truth interestingness value. (left)
X-axis: each dimension; Y-axis: Correlation values; (right): histogram
of the correlation for different features.}
}}
\end{figure*}

\textit{\textcolor{blue}{}}
\begin{table*}
\begin{centering}
{\tiny{}}%
\begin{tabular}{|c||c|c|c|c|c|c|c|c|}
\hline 
{\tiny{}attribute} & {\tiny{}pleasant\_scene} & {\tiny{}attractive} & {\tiny{}memorable} & {\tiny{}is\_aesthetic} & {\tiny{}is\_interesting} & {\tiny{}on\_post-card} & {\tiny{}buy\_painting } & {\tiny{}hang\_on\_wall }\tabularnewline
\hline 
\hline 
{\tiny{}corr} & {\tiny{}-0.4060 } & {\tiny{}-0.4273 } & {\tiny{}-0.4618 } & {\tiny{}0.4487 } & {\tiny{}0.4715 } & {\tiny{}0.4767} & {\tiny{}0.4085} & {\tiny{} 0.4209}\tabularnewline
\hline 
\end{tabular}
\par\end{centering}{\tiny \par}

\textit{\textcolor{blue}{\protect\caption{\label{tab:The-pruned-attribute}The pruned attribute features.}
}}
\end{table*}

\textit{\textcolor{blue}{So as suggested, for more fair comparisons,
we remove the attribute features \cite{Isola2011NIPS} whose kendall
tau correlations are higher than 0.4 or lower than -0.4. This will
lead to deletion the features listed in Table \ref{tab:The-pruned-attribute}.
These pruned features include those suggested by Reviewer 3 (``is\_interesting''
and ``is aesthetic''). but not the ``unusual'' attribute feature
which has a low correlation value of -0.0226. }}

\textit{\textcolor{blue}{We repeat the image interestingness experiments
with the updated features. It is noticed that this has little effect
on the results (still within the variances). }}\textit{\textcolor{red}{}}

\end{enumerate}

\end{document}